\documentclass[letterpaper, conference]{IEEEtran}
\IEEEoverridecommandlockouts

\usepackage{color,soul}
\usepackage{amsmath}
\usepackage{siunitx}
\usepackage{url}

\ifCLASSINFOpdf
  \usepackage[pdftex]{graphicx}
  \graphicspath{{../pdf/}{../figures/}}
  \DeclareGraphicsExtensions{.pdf,.jpeg,.png}
\else
  \usepackage[dvips]{graphicx}
  \graphicspath{{../eps/}}
  \DeclareGraphicsExtensions{.eps}
\fi

\ifCLASSOPTIONcompsoc
  \usepackage[caption=false,font=normalsize,labelfont=sf,textfont=sf]{subfig}
\else
  \usepackage[caption=false,font=footnotesize]{subfig}
\fi

\usepackage{tikz}

\newcommand\copyrightnotice{%
\begin{tikzpicture}[remember picture,overlay]
\node[anchor=south,yshift=10pt] at (current page.south) {\fbox{\parbox{\dimexpr\textwidth-\fboxsep-\fboxrule\relax}{\footnotesize \textcopyright 2017 IEEE. Personal use of this material is permitted. Permission from IEEE must be obtained for all other uses, in any current or future media, including reprinting/republishing this material for advertising or promotional purposes, creating new collective works, for resale or redistribution to servers or lists, or reuse of any copyrighted component of this work in other works.}}};
\end{tikzpicture}%
}

\begin{document}
%
\title{Automatic Extrinsic Calibration for Lidar-Stereo Vehicle Sensor Setups
\\[-0.5ex]}

\author{\IEEEauthorblockN{Carlos Guindel\IEEEauthorrefmark{1},
Jorge Beltr\'{a}n\IEEEauthorrefmark{1}, David Mart\'{i}n, \IEEEmembership{Member, IEEE}, 
and Fernando Garc\'{i}a, \IEEEmembership{Member, IEEE}
}
\IEEEauthorblockA{Intelligent Systems Laboratory (LSI) Research Group\\
Universidad Carlos III de Madrid, Legan\'{e}s, Madrid, Spain\\
\{cguindel, jbeltran, dmgomez,
fegarcia\}@ing.uc3m.es}
\thanks{\IEEEauthorrefmark{1}The first two authors contributed equally to this work.}
\\[-6ex]
}

\maketitle
\copyrightnotice

\begin{abstract}
Sensor setups consisting of a combination of 3D range scanner lasers and stereo vision systems are becoming a popular choice for on-board perception systems in vehicles; however, the combined use of both sources of information implies a tedious calibration process. We present a method for extrinsic calibration of lidar-stereo camera pairs without user intervention. Our calibration approach is aimed to cope with the constraints commonly found in automotive setups, such as low-resolution and specific sensor poses. To demonstrate the performance of our method, we also introduce a novel approach for the quantitative assessment of the calibration results, based on a simulation environment. Tests using real devices have been conducted as well, proving the usability of the system and the improvement over the existing approaches. Code is available at \url{http://wiki.ros.org/velo2cam_calibration}.
\end{abstract}


%
\IEEEpeerreviewmaketitle

\section{Introduction}
Advanced Driver Assistance Systems (ADAS) and upcoming autonomous cars rely on an accurate perception of the environment to make the proper decisions concerning the trajectory of the vehicle. These systems must be endowed with an exceptional degree of robustness under different circumstances that can affect the driving decisions (e.g. illumination).

Consequently, the design of perception systems intended for automotive applications is usually oriented towards topologies with several complementary sensory modalities. Vision systems are frequent \cite{Franke2013}, due mainly to their ability to provide appearance information. Among the available vision devices, stereo-vision systems, which makes use of a pair of cameras separated a fixed distance to get depth information about the environment, stand out as a cost-effective solution able to provide additional dense 3D information to model the surroundings of the vehicle. 

On the other hand, the rapid development of the technology behind 3D laser scanners in recent years has enabled its widespread use in both research and industry driving applications. Contrary to vision systems, lidar range measurements are characterized by its high accuracy; moreover, they can provide information in a full 360\si{\degree} field of view. 

\begin{figure}[t]
\centering
\subfloat[]{\includegraphics[height=1.2in]{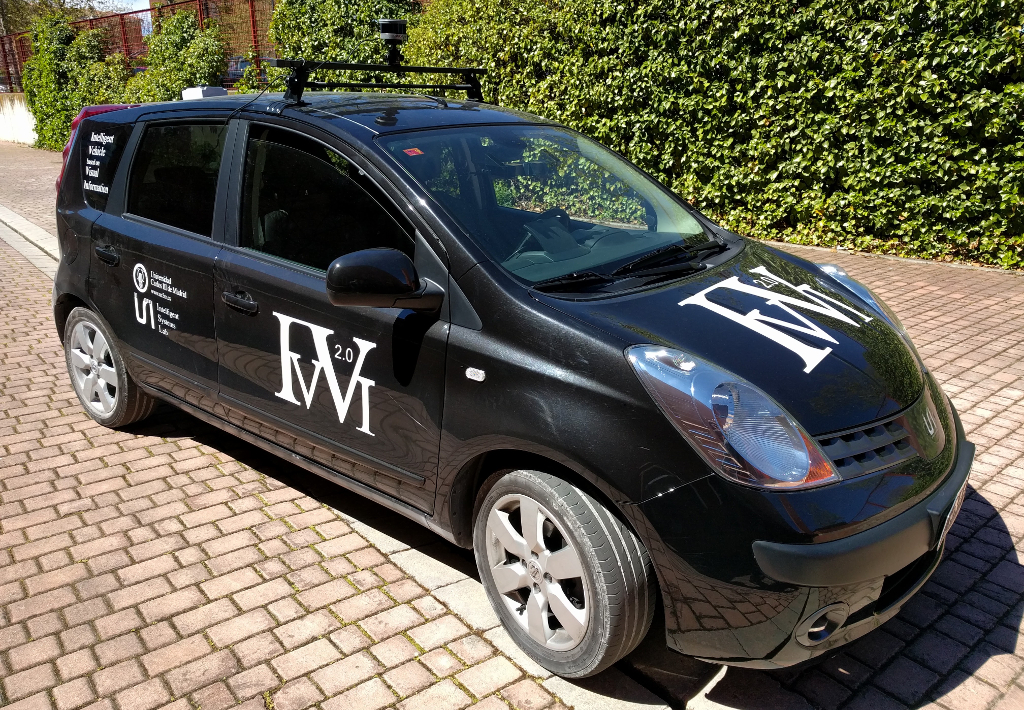}%
\label{fig:ivvi_outside}}
\hfil
\subfloat[]{\includegraphics[height=1.2in]{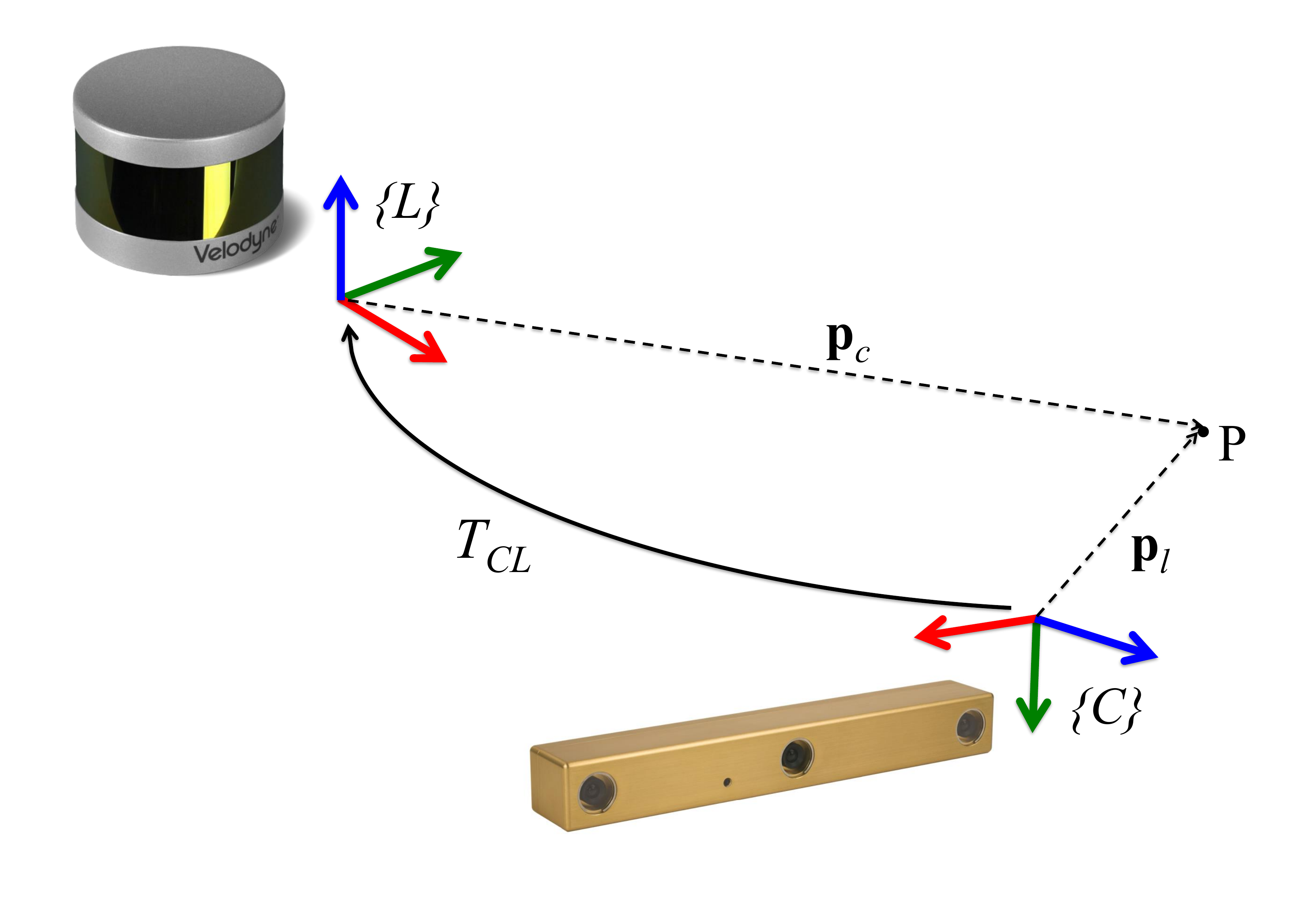}%
\label{fig:sensor_scheme}}
\caption{(a) Common sensor setup for vehicles using a laser scanner and a trinocular camera; (b) relative calibration between both sensors}
\label{fig:sidesetup}
\end{figure}

Due to the particular features of these two sensory technologies, they are suitable to be part of the same perception system, providing complementary information. In that kind of designs, data from both sensors have to be appropriately combined before inference, making use of fusion techniques \cite{Garcia2017}. In the most usual setup, sensors have overlapping fields of view (as in Fig. \ref{fig:sidesetup}), and the advantages conferred by their joint use come from the ability to make correspondences between both data representations. For that end, the relative pose of the sensors, given by their respective extrinsic parameters, must be accurately determined through a calibration process.

Existing calibration methods suffer from either the need for complex setups or the lack of generalization ability, so that the accuracy of the results is strongly dependent on the parameters of the sensors or the structuredness of the environment. 

In this work, we present a calibration method tailored to automotive sensor setups composed of a stereo rig and a 360\si{\degree} multi-layer lidar scanner. Unlike the existing methods, no strong assumptions are made, allowing its use with medium-resolution scanners (e.g. 16-layers) as well as very different relative poses between the sensors. Our method can be performed within a reasonable time using a simple setup designed to exploit the correspondences in the data from both devices. 

The remainder of this paper is organized as follows. In Section \ref{sec:related}, a brief review of related work is provided. Section \ref{sec:proposed} details a description of the proposed algorithm. In Section \ref{sec:results}, the experimental results that assess the method performance are discussed. Finally, in Section \ref{sec:conclusion}, conclusions are presented.
	
\section{Related work}
\label{sec:related}
The issue of calibration of the extrinsic parameters relating different sensors has been addressed by many researchers in the past, driven by its typical application in multi-sensory robotic and automotive platforms. 

Among the variety of possible sensor setups, the camera-to-range problem is the most frequently considered in the literature. Due to the restrictions inherent to mobile platforms, devices used in this applications often provide data in a limited number of scan planes, typically one or four, resulting in very scarce range information \cite{Kwak2011}. Calibration is usually assumed as a process to be performed in a controlled environment before the regular operation of the sensorial system. Traditional methods require manual annotation to some extent \cite{Scaramuzza2007}. However, since miscalibrations are frequent in robotic platforms, research effort has usually focused on automatic approaches. As the process aims to find the correspondence between data acquired from different points of view, unambiguous pattern instruments have been used as calibration targets, such as triangular boards \cite{Debattisti2013}, polygonal boards \cite{Park2014} or spheres \cite{Pereira2016}. The diversity of shapes used in the existing works deals with the necessity of the targets to be distinguishable in all the data representations from the sensors (i.e. range and appearance). Nonetheless, planar targets are particularly prevalent \cite{Li2011}, since they are easily detectable using range information and provide a characteristic shape that can be used to perform geometrical calculations. On the other hand, Scott et al. dealt with the particular case of non-overlapping field of views \cite{Scott2015} using an accurate motion tracking system. 

With the widespread introduction of range sensors providing a dense 3D point cloud in recent years, research interest has shifted to these type of devices. Geiger et al. \cite{Geiger2012c} proposed a calibration method based on a single shot in the presence of a setup based on several planar checkerboards used as calibration targets. Velas et al. \cite{Velas2014} propose an approach enabling the estimation of the extrinsic parameters using a single point of view, based on the detection of circular features on a calibration pattern. These methods are targeted to dense range measurements so that 3D lidar scanners with lower resolution (e.g. the 16-layer scanner used in this work) entail particular issues that are addressed in this paper. 

A large second group of approaches dispenses with any artificial calibration targets and use the features in the environment. Moghadam et al. \cite{Moghadam2013} use linear features extracted from natural scenes to determine the transformation between the coordinate frames. The method is suitable for indoor scenes populated with a large number of linear landmarks. In traffic environments, the ground plane and the obstacles have been used to perform camera-laser calibration \cite{Ponz}, although some parameters are assumed as known. 

On the other hand, the assessment of calibration methods remains an open issue, given that an accurate ground-truth of the six parameters defining the relationship between the pose of the sensors cannot be obtained in practice. The lack of standard evaluation metrics has led to the use of custom schemes, which are difficult to extend to other domains and eventually based on inaccurate manual annotations. In this regard, Levinson and Thrun \cite{Levinson2013} presented a method aimed to detect miscalibrations through the variations in an objective function computed from the discontinuities in the scene. 

\section{Calibration Algorithm}
\label{sec:proposed}
We propose a calibration algorithm aimed to estimate the rigid-body transform relating the coordinate system $\left\{ C \right\}$, centered in one of the two cameras belonging to the stereo-vision system (hereafter assumed to be the leftmost one), and the coordinate frame $\left\{ L \right\}$, fixed in the laser range scanner, as depicted in Fig. \ref{fig:sensor_scheme}. This transformation is defined by a set of six parameters $\boldsymbol{\xi}_{CL} = (t_x, t_y, t_z, \phi, \theta, \psi)$ representing the translation along the $x$, $y$ and $z$ axis, and the rotation around $x$ (roll), $y$ (pitch) and $z$ (yaw). We compute the parameters with respect to the stereo camera frame $\left\{ C \right\}$, although the transformation is straightforwardly reversible. Using homogeneous coordinates, the set $\boldsymbol{\xi}_{CL}$ can be used to build a transformation matrix, $\mathbf{T}_{CL}$, which allows to transform a 3D point in camera coordinates, $\mathbf{p}_c$, into a point in laser range scanner coordinates, $\mathbf{p}_l = \mathbf{T}_{CL}\mathbf{p}_c$.

To obtain this transform, we use a short series of stereo-pair images (where both images are expected to be synchronized) and lidar scans. We assume that the intrinsic parameters relating the 3D scene points with the representations provided by both sensors are known, including the baseline of the stereo system, whose images are expected to be rectified beforehand.  

\begin{figure}[htb]
\centering
\subfloat[]{\includegraphics[height=1.2in]{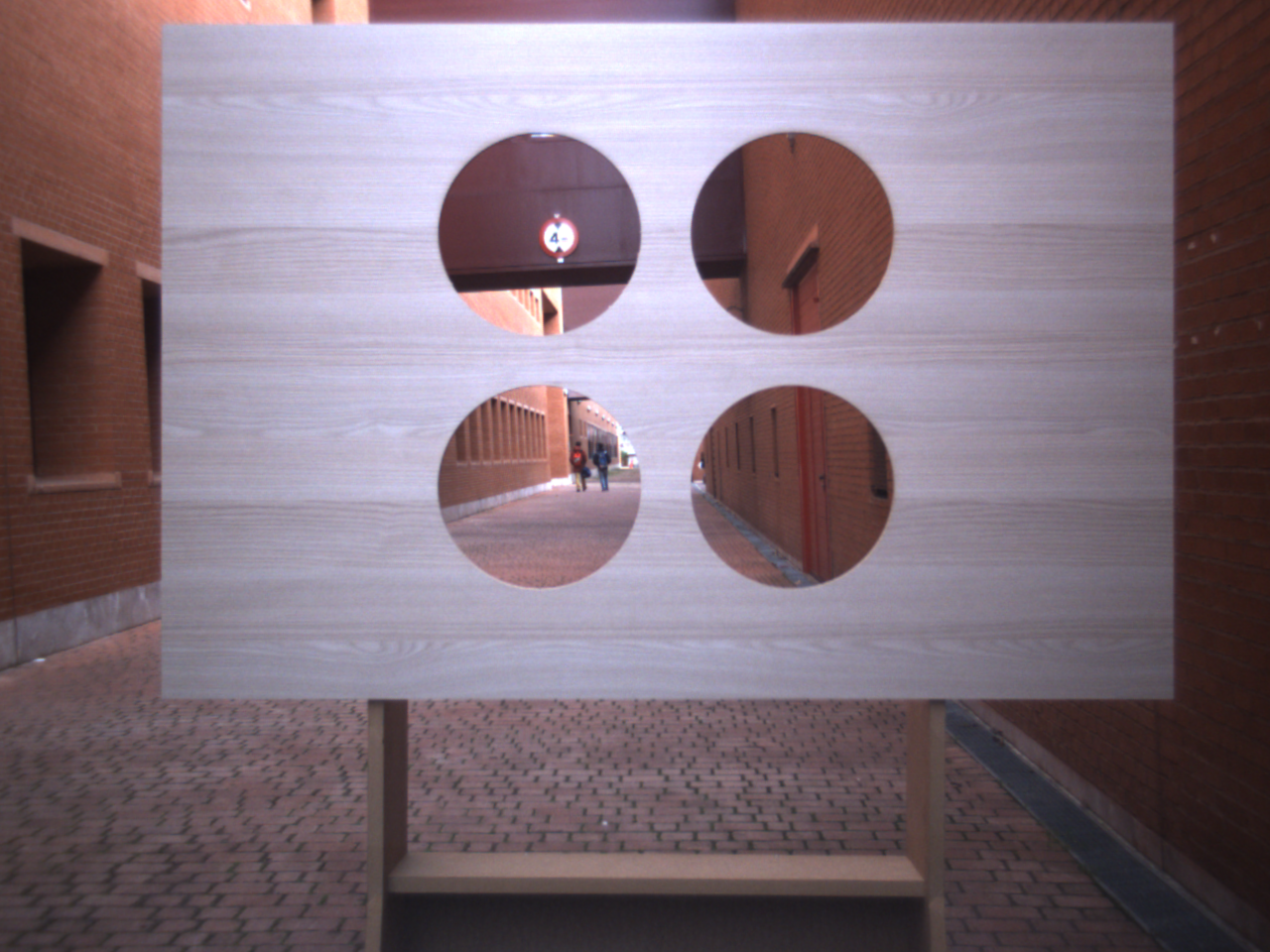}%
\label{fig:patron_img}}
\hfil
\subfloat[]{\includegraphics[height=1.2in]{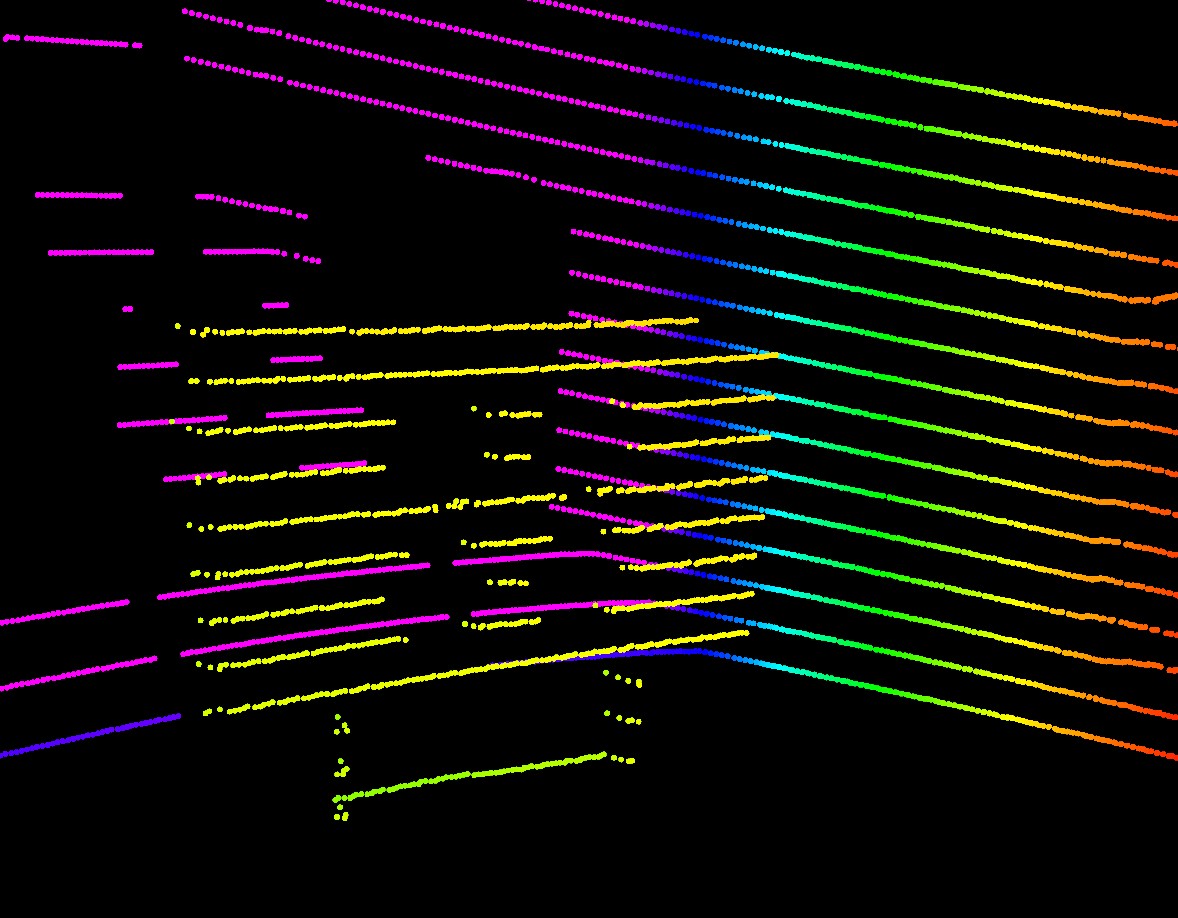}%
\label{fig:patron_pc}}
\caption{(a) Calibration target for the proposed method, as seen in the left image of the stereo system; (b) projection in a 16-layer laser point cloud.}
\label{fig:escobar}
\end{figure}

A single calibration target is used, and it is intended to be perceived by the sensors from a unique point of view, avoiding the need for the presence of multiple targets or changes in the calibration scene during the process. Following \cite{Velas2014}, we use a custom-made planar target, with four circular holes symmetrically disposed, as shown in Fig. \ref{fig:escobar}. Those holes act as distinct features visible by the camera and the lidar; therefore, both sensors are assumed to be placed with a certain overlap in their field of views, so that the circular holes are intersected by at least two lidar beams and fully visible from the camera.

Beyond that reasonable constraint, no additional assumptions are made regarding either the relative rotation and translation of the sensors (i.e. large displacements are allowed) or the pose of the calibration target, which is not required to be perfectly aligned with any axis. 

\begin{figure*}[t]
\centering
\subfloat[]{\includegraphics[height=1in]{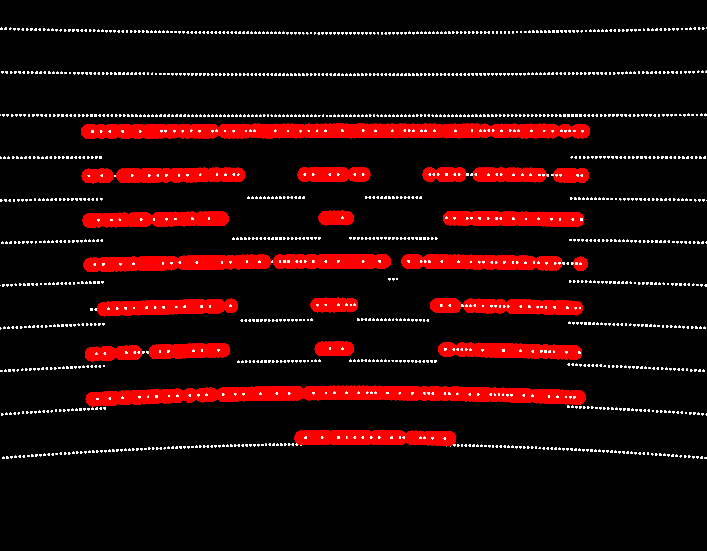}%
\label{fig:full1l}}
\hfil
\subfloat[]{\includegraphics[height=1in]{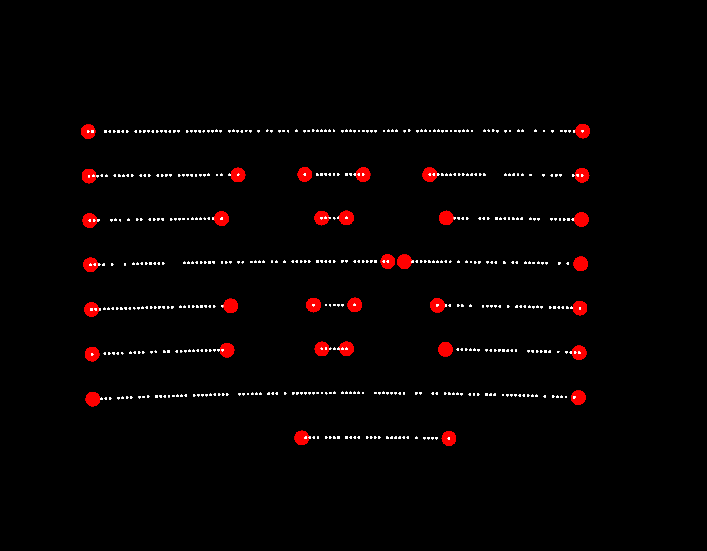}%
\label{fig:full2l}}
\hfil
\subfloat[]{\includegraphics[height=1in]{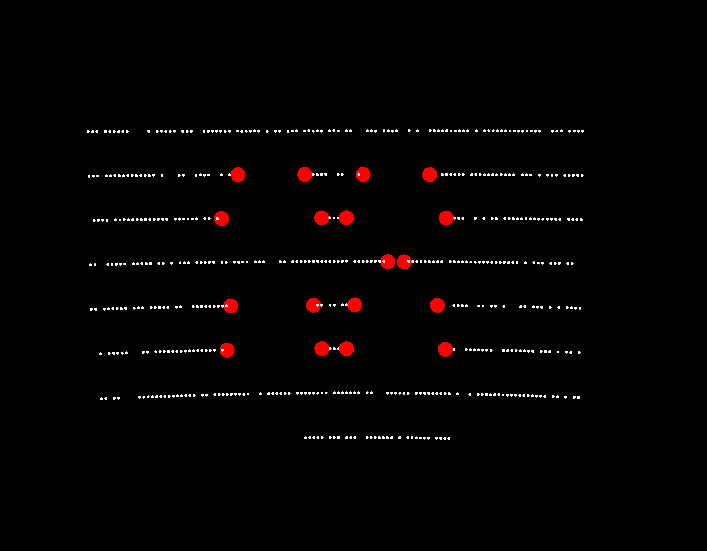}%
\label{fig:full3l}}
\hfil
\subfloat[]{\includegraphics[height=1in]{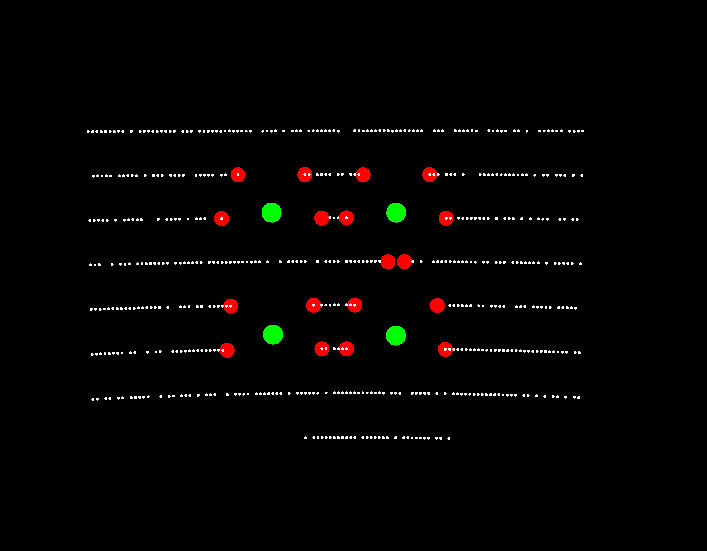}%
\label{fig:full4l}}
\hfil
\subfloat[]{\includegraphics[height=1in]{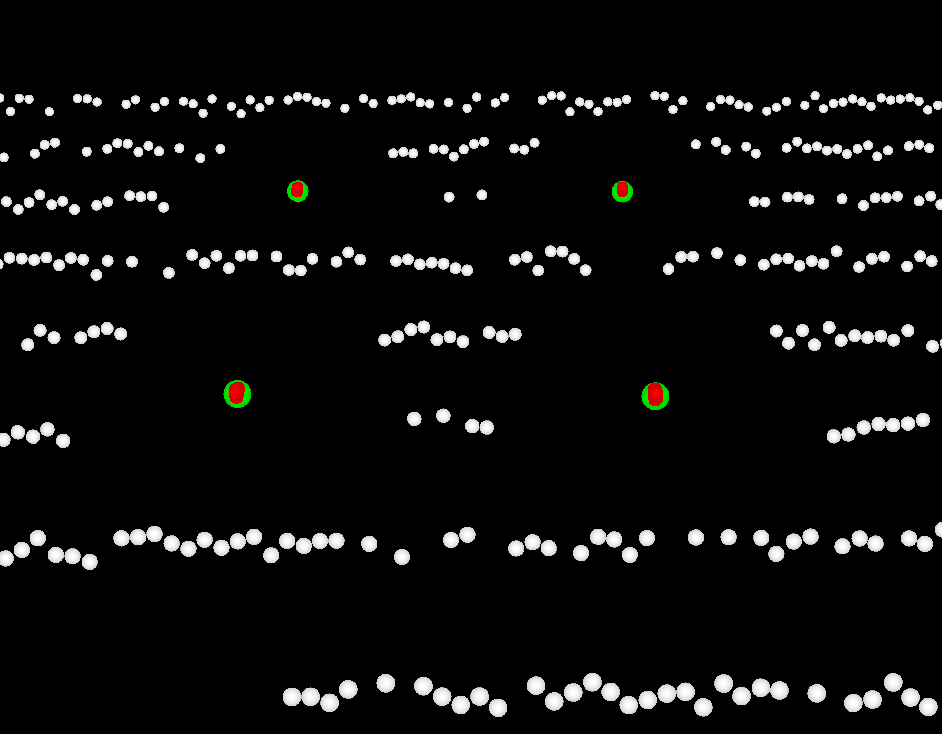}%
\label{fig:full5l}}
\caption{Segmentation pipeline for extraction of the reference points from the lidar point cloud.}
\label{fig:fulll}
\end{figure*}

\begin{figure*}[t]
\centering
\subfloat[]{\includegraphics[height=1in]{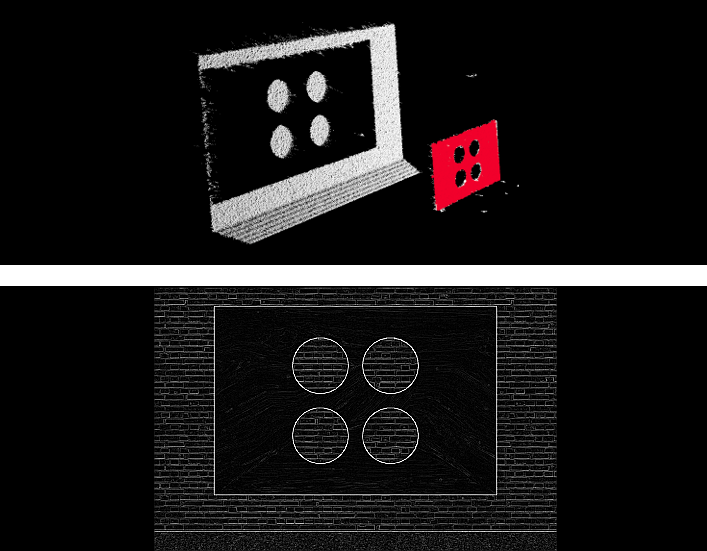}%
\label{fig:full1c}}
\hfil
\subfloat[]{\includegraphics[height=1in]{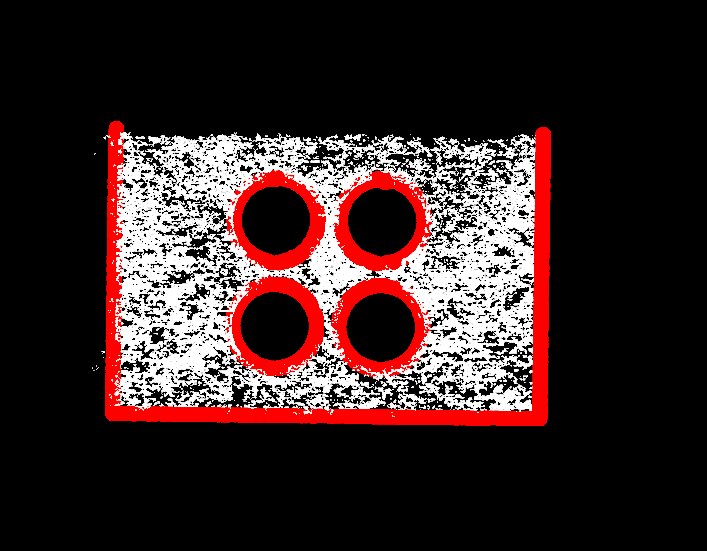}%
\label{fig:full2c}}
\hfil
\subfloat[]{\includegraphics[height=1in]{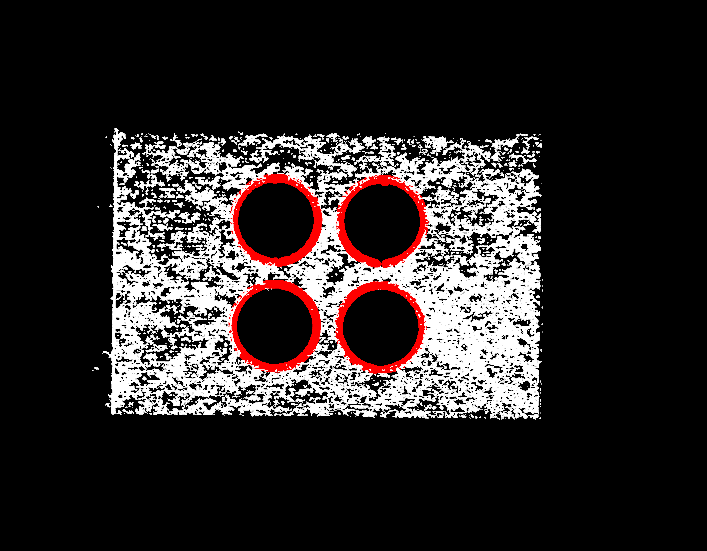}%
\label{fig:full3c}}
\hfil
\subfloat[]{\includegraphics[height=1in]{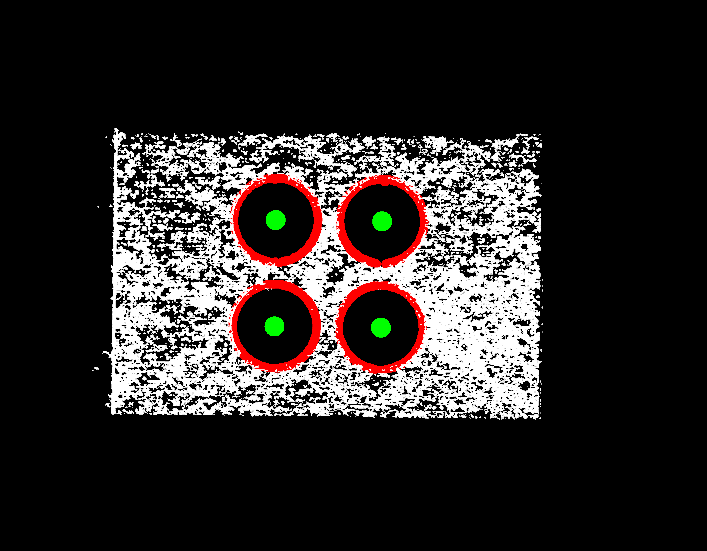}%
\label{fig:full4c}}
\hfil
\subfloat[]{\includegraphics[height=1in]{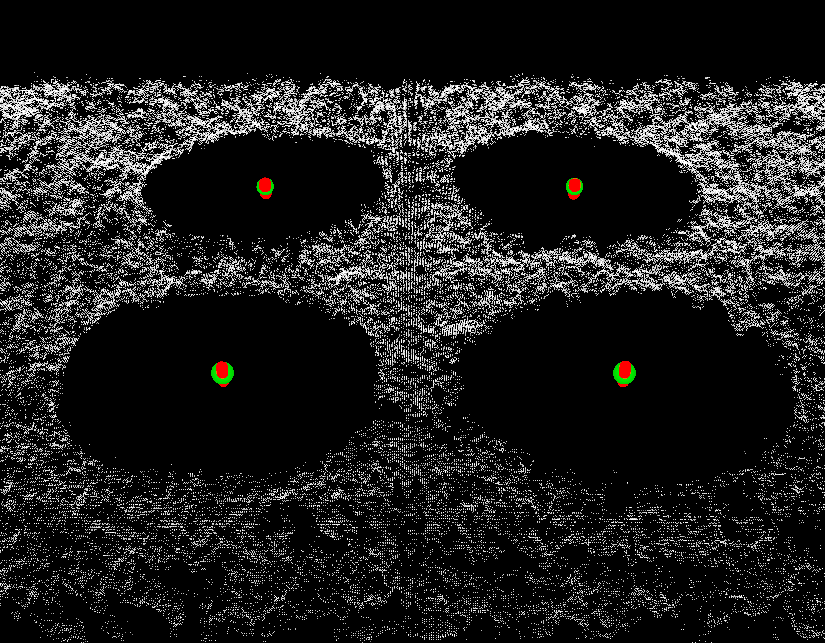}%
\label{fig:full5c}}
\caption{Segmentation pipeline for extraction of the reference points from the stereo point cloud.}
\label{fig:fullc}
\end{figure*}

The calibration process involves two stages: a segmentation of reference points in both clouds, illustrated in Fig. \ref{fig:fulll} and Fig. \ref{fig:fullc}, and a registration to estimate the transform parameters.


\subsection{Data Representation}
The proposed method rely on the information provided by the sensors to be calibrated. For the laser scanner, the 3D point cloud $\mathcal{P}^l_0 = \left\{(x,y,z)\right\}$ representing the range measurements is considered. On the other hand, information from the stereo camera is employed in its two modalities: grayscale intensity and depth estimation. The latter is obtained by a stereo matching algorithm, which allows to map every pixel to its $(x,y,z)$ coordinates, thus producing an analogous point cloud $\mathcal{P}^c_0 = \left\{(x,y,z)\right\}$. The similarity between both clouds is exploited during the calibration process for, ultimately, determining the transformation between both sensors. Each of the clouds $\mathcal{P}^l_0$ and $\mathcal{P}^c_0$ is expressed in the coordinate system with origin in its sensor; that is, $\left\{ L \right\}$ and $\left\{ C \right\}$ respectively.

For the stereo matching stage, we use the Semi-Global Matching (SGM) method \cite{Hirschmuller2008}, which we found to be reasonably accurate in the depth estimation. The border localization problem typically present in stereo matching do not significantly affect the algorithm, since it is tackled by using the intensity information, as will be shown below. However, the calibration target is assumed to present a minimum of texture, allowing the resolution of the stereo correspondence problem.

Pass through filters, based on the distance along the local axes, are then applied to the point clouds from the two sensors. This step is useful to limit the information processing to the area which is likely covered by both fields of views, where the calibration target must be placed.

\subsection{Target segmentation}
The first steps of the calibration algorithm are intended to extract the points in each cloud belonging to discontinuities in the calibration target. For this purpose, several successive segmentation processes are carried out over the clouds containing the data from the sensors. Segmentation is used to find subsets of points $\mathcal{P}_{i_0}$ representing a geometrical shape (e.g. a plane) in the original cloud. Thus, for the step $i_0$:
\begin{equation}
\mathcal{P}_{i_0}^j = \left\{(x,y,z)\right\} \subseteq \mathcal{P}_{i_0-1}^j, \;\; \forall j \in \left\{ l,c \right\}
\end{equation}

Taking advantage of the planar shape of the calibration target, we perform a sample consensus-based plane segmentation method to determine the plane models in each cloud: $\pi^c$ and $\pi^l$. Some restrictions are posed to guarantee the segmentation of the proper plane in different types of real environments, including those where the ground plane or building walls could affect the segmentation. First, a tight threshold $\delta_{plane}$ is used in the RANSAC algorithm and, second, the plane model is required to be parallel to the vertical axis of the sensor reference frame within a specified angular deviation $\alpha_{plane}$.

When the plane model is available, points separated from the model a distance greater than $\delta_{inliers,l}$, for the lidar cloud, and $\delta_{inliers,c}$, for the stereo cloud, are removed, resulting in the cloud segments $\mathcal{P}_1^l$ and $\mathcal{P}_1^c$. Examples of the results of the plane segmentation process are shown in Fig. \ref{fig:full1l} and Fig. \ref{fig:full1c}.

After that, the segmented clouds undergo a process aimed to filter out the points in the calibration target not belonging to discontinuities. Due to the nature of the data from the two sensors, this stage is sensor-specific. 

Regarding the laser point cloud, we follow the method in \cite{Levinson2013} to find depth discontinuities. Each point in the plane model cloud, $\mathbf{p}^i \in \mathcal{P}_1^l$, is assigned a magnitude representing the depth difference with respect to their neighbors:
\begin{equation}
p^i_\Delta = \max(p^{i-1}_r-p^{i}_r,p^{i+1}_r-p^{i}_r,0)
\end{equation}
Where $p_r^i$ is the range measurement given by the sensor for the point $\mathbf{p}^i$, and  $\mathbf{p}^{i-1}$ and $\mathbf{p}^{i+1}$ are the points adjacent to $\mathbf{p}^{i}$ in the same scan plane. Then, we filter out all points with a discontinuity value $p_\Delta < \delta_{discont,l}$ (Fig. \ref{fig:full2l}). 

On the other hand, for the stereo-camera cloud, we keep the points that map to edges in the intensity image. To that end, a Sobel filter is applied over the left image of the stereo pair. Points whose projection on the image corresponds to a low value in the Sobel image (smaller than $\tau_{sobel,c}$) are filtered out, as shown in Fig. \ref{fig:full2c}.

\subsection{Circle segmentation}
Next steps are intended to segment the four circles on the calibration target, whose centers will be used as correspondence keypoints between the two clouds in the registration. 

To enhance the circle segmentation accuracy, we previously perform a filtering process aimed to get rid of the points not belonging to the circles. As only discontinuities are present in the clouds at this stage, outliers to be removed are, mostly, from the outer boundaries of the calibration target. The cloud from the lidar is processed to keep only the rings with a number of points compatible with the presence of a circle and subsequently to remove the outer points in these rings. On the other hand, the camera cloud is subjected to a filtering process based on the elimination of lines, since the borders of the calibration target are densely represented in this case. Lines are found using a sample consensus segmentation and selected according to their orientation and the known dimensions of the calibration pattern, to prevent the removal of useful information from the circles themselves. Notwithstanding these considerations, our experiments proved that our segmentation method was largely insensitive to the presence of these borders, except for some very specific poses of the calibration target. Filtered clouds, $\mathcal{P}_2^l$ and $\mathcal{P}_2^c$, are shown in Fig. \ref{fig:full3l} and Fig. \ref{fig:full3c}.

Afterward, points representing the holes of the calibration target in both clouds are detected. To that end, a circle segmentation process is performed in the 2D space determined by the plane models $\pi^c$ and $\pi^l$. This is effectively implemented by rotating the clouds $\mathcal{P}_2$ until the points are aligned with the XY plane and then adjusting their $z$ coordinate to that enforced by the plane equation. Later, circles are segmented in the XY subspace through sample consensus, imposing the known circle radius as a constraint. To avoid spurious detections, the distance between the circle centers in the physical calibration target is also taken into account. Finally, the obtained centers are transformed back to the 3D sensor's coordinate frame, resulting in the point clouds depicted in green in Fig. \ref{fig:full4l} and Fig. \ref{fig:full4c}. Note that, since the circle segmentation stage is performed in a bidimensional space, the proposed method is suitable for medium-resolution range sensors, as circles are defined by only three points in a plane, thus requiring just two lidar beams to intersect with each of the four circles.

Provided that the representation of the calibration target is accurate enough in the data from both sensors, registration might be performed with a single-shot representation of the four circles centers. Nevertheless, in order to improve the robustness of the method against the sources of noise present in the process, centers are cumulated over a window of $N$ frames. Later, a clustering algorithm is applied, and the cluster centroids are used at the registration stage. This implies the assumption that the calibration target, as well as the sensors, remains static throughout the whole process. We choose a Euclidean distance strategy, with a cluster tolerance of $\delta_{cluster}$, to perform the clustering so that this approach would naturally tackle the presence of outliers. Strict restrictions can be imposed on the minimum and the maximum number of points allowed in each cluster considering the length of the window, i.e. the number of frames, used in the computation.

\subsection{Registration}
The final registration process is aimed to find the set of transformation parameters $\hat{\boldsymbol{\xi}}_{CL}$ which minimize the distance between the reference points of both clouds, namely the centroids of the cumulated circle centers, once the transformation is applied. 

The registration procedure involves two steps. First, we compute the optimal transformation assuming the absence of rotation. The result is hence a pure translation, expressed by the set of parameters $\mathbf{t}_{CL}' = (t_x', t_y', t_z')$, which can be obtained by finding the least-squares solution of the overdetermined system of 12 equations provided by the registration of the four reference points:
\begin{equation}
\mathbf{t}_{CL} = \bar{\mathbf{p}}^i_l-\bar{\mathbf{p}}^i_c, \;\; \forall i \in \left\{tl, tr, bl, br \right\}
\end{equation}
Three equations are obtained for each pair of centroids from both the laser cloud, $\bar{\mathbf{p}}^i_l$, and the stereo camera cloud, $\bar{\mathbf{p}}^i_c$. Points are labeled according to its position in the sensors' coordinate systems as top-left ($tl$), top-right ($tr$), bottom-left ($bl$) and bottom-right ($br$) to allow the matching of the points in both clouds. Finally, we find the least-squares solution through column-pivoting QR decomposition. 

In the second stage of the registration process, we provide a final estimation of the full set of parameters of the transform: $\hat{\boldsymbol{\xi}}_{CL} = (t_x, t_y, t_z, \phi, \theta, \psi)$. Given that the data association problem is trivial to solve at this stage, we rely on the well-known Iterative Closest Points algorithm \cite{Besl1992} to minimize the sum of point-to-point distances between the cluster centroids clouds. The final transformation is straightforwardly obtained as the composition of that two partial transformations. 

\section{Experimental Results}
\label{sec:results}

The performance of the proposed calibration approach is evaluated in a comprehensive set of experiments. The analysis includes both tests performed in a synthetic test suite developed for the evaluation of calibration algorithms, and real tests with the sensing devices available in the IVVI 2.0 research platform \cite{Martin2014a} (shown in Fig. \ref{fig:ivvi_outside}). The former are used to provide numerical results against a perfect ground-truth; moreover, they allow comparison with other approaches in the literature. Meanwhile, the later demonstrate the validity of the method in real world conditions. 

\subsection{Synthetic Test Suite}

Unlike previous works, where calibration performance is compared to manual annotations \cite{Geiger2012c} or scene discontinuities \cite{Levinson2013}, we propose a novel approach based on a simulation environment, which provides exact ground truth about the relative transformation between sensors. To this end, the open-source Gazebo simulator \cite{Koenig2004} is used, and nine different calibration setups are defined, as listed in Table \ref{tab:setups}. The first seven settings are simple setups designed so that the parameters of the transform could be independently evaluated, while the last two represent challenging situations that go beyond the difficulties that may arise in automotive applications, in order to analyze the ability of generalization of the proposed approach. 
\begin{table}[hbt]
    \renewcommand{\arraystretch}{0.9}
    \caption{Camera-lidar transformation parameters in the simulator settings used for the experiments}
    \label{tab:setups}
    \centering
    \begin{tabular}{| c | c c c | c c c |}
    \hline
    Setting & $t_x$ (m) & $t_y$ (m) & $t_z$ (m) & $\psi$ (rad) & $\theta$ (rad) & $\phi$ (rad)\\ \hline
    1 & -0.8 &    -0.1    &    0.4    &    0    &    0    &    0 \\ 
    2 & 0     &    0        &      0    &    0.5    &    0    &    0 \\
    3 & 0     &    0        &    0    &    0.3    &    0.1    &    0.2 \\
    4 & -0.3 &    0.2        &    -0.2&    0.3    &    -0.1&    0.2\\ 
    5 & 0     &    0        &      0    &    0    &    0.1    &    0 \\
    6 & 0     &    0        &    0    &    0    &    0    &    0.4 \\
    7 & 0     &    0        &     0    &    0    &    0    &    0 \\ 
    8 & -0.128    & 0.418 & -0.314 & -0.103 &    -0.299 & 0.110 \\
    9 & -0.433    & 0.845 & 1.108 & -0.672 & 0.258 & 0.075 \\
    \hline
    \end{tabular}
\end{table}

The stereo camera and different lidar devices were modeled taking into account their real specifications in terms of field of view, resolution, and accuracy. As a consequence, calibration results achieved in the calibration scenarios are comparable to those obtained in real settings. A model of the calibration target was created mimicking the appearance of the actual wooden embodiment shown in Fig. \ref{fig:patron_img}, so that realism was preserved.

\subsection{Performance Evaluation}
Algorithm parameters were empirically chosen as: $\delta_{plane}$ = \SI{1}{cm}, $\delta_{inliers,l}$ = \SI{5}{cm}, $\delta_{inliers,c}$ = \SI{10}{cm}, $\delta_{discont,l}$ = \SI{50}{cm}, $\delta_{cluster}$ = \SI{2}{cm}, $\tau_{sobel,c}$ = 128 and $\alpha_{plane}$ = 0.55 rad.

Following \cite{Geiger2012c}, performance was evaluated as the difference between the estimated transform and the ground-truth, measured in its linear and angular components:
\begin{align}
e_t = \| \mathbf{t}-\mathbf{t_g} \| \\
e_r = \angle(\mathbf{R}^{−1}\mathbf{R}_g)
\end{align}

Where $\mathbf{t}$ is the translation vector given by the $\tau_{CL}$ parameters and $\mathbf{R}$ is the rotation matrix provided by the roll, pitch and yaw angles. 

The baseline model was based on a Point Grey Bumblebee XB3 trinocular system (resolution 1280x960, baseline \SI{12}{cm}) and a 16-layer Velodyne VLP-16 lidar scanner. Unless stated otherwise, three different results are given for each setting. A Gaussian noise $\mathcal{N}(0, \sigma_0^2)$ was included in the measurements of the sensors, with $\sigma_{0}^c=0.007m$ for the pixel intensities and $\sigma_{0}^l=0.008m$ for the range measurements from the lidar.

\begin{figure}[htb]
\centering
\subfloat[Translation error, $e_t$]{\includegraphics[height=1.12in]{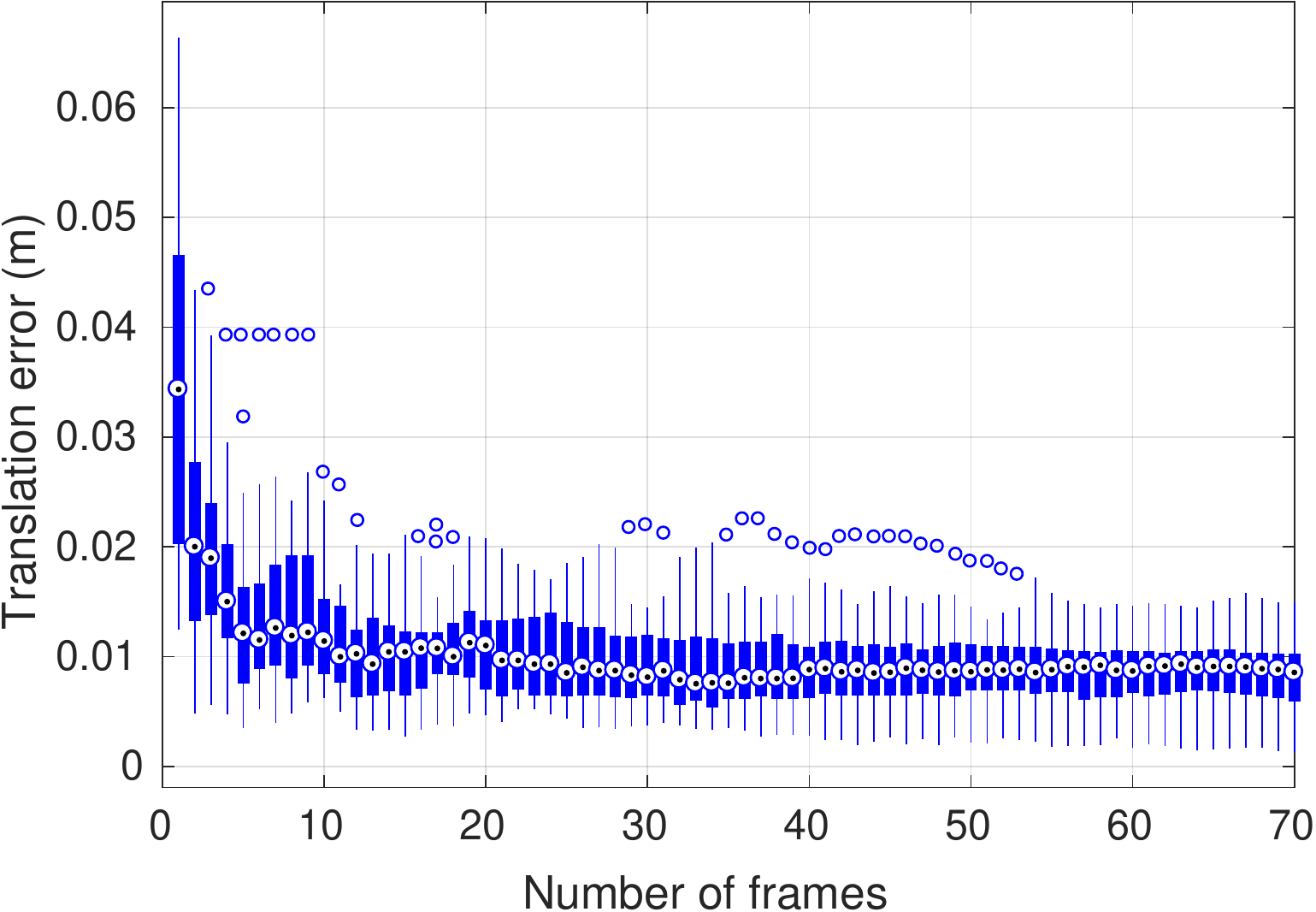}%
\label{fig:tr_iter}}
\hfil
\subfloat[Rotation error, $e_r$]{\includegraphics[height=1.12in]{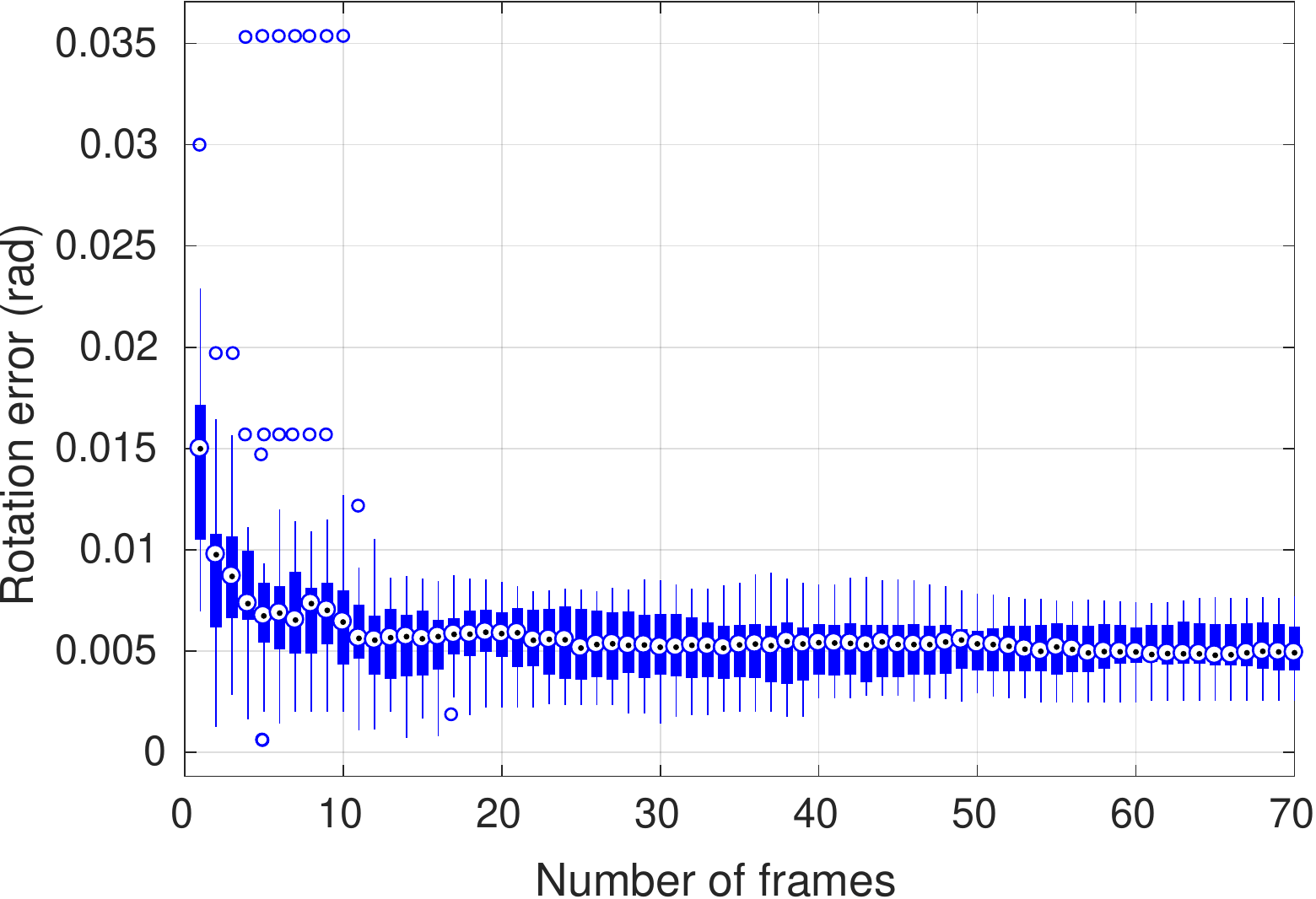}%
\label{fig:or_iter}}
\caption{Evolution of the calibration errors with the number of cumulated frames. Henceforth, boxes represents the interquartile range, and outliers are depicted individually. }
\label{fig:iter}
\end{figure}

To test the sensitivity of the algorithm to the length of the window $N$, the error after each number of data shots provided by the sensors is plotted in Fig. \ref{fig:iter}. Both the linear and the angular error converge quickly, showing the accuracy of the proposed method. From now on, results are given for the 30\textsuperscript{th} iteration, where the error rate becomes steady. Note that not every frame is actually used for the calibration since the segmentation process is not guaranteed to provide an outcome. 

\begin{figure}[htb]
\centering
\subfloat[Translation error, $e_t$]{\includegraphics[height=1.12in]{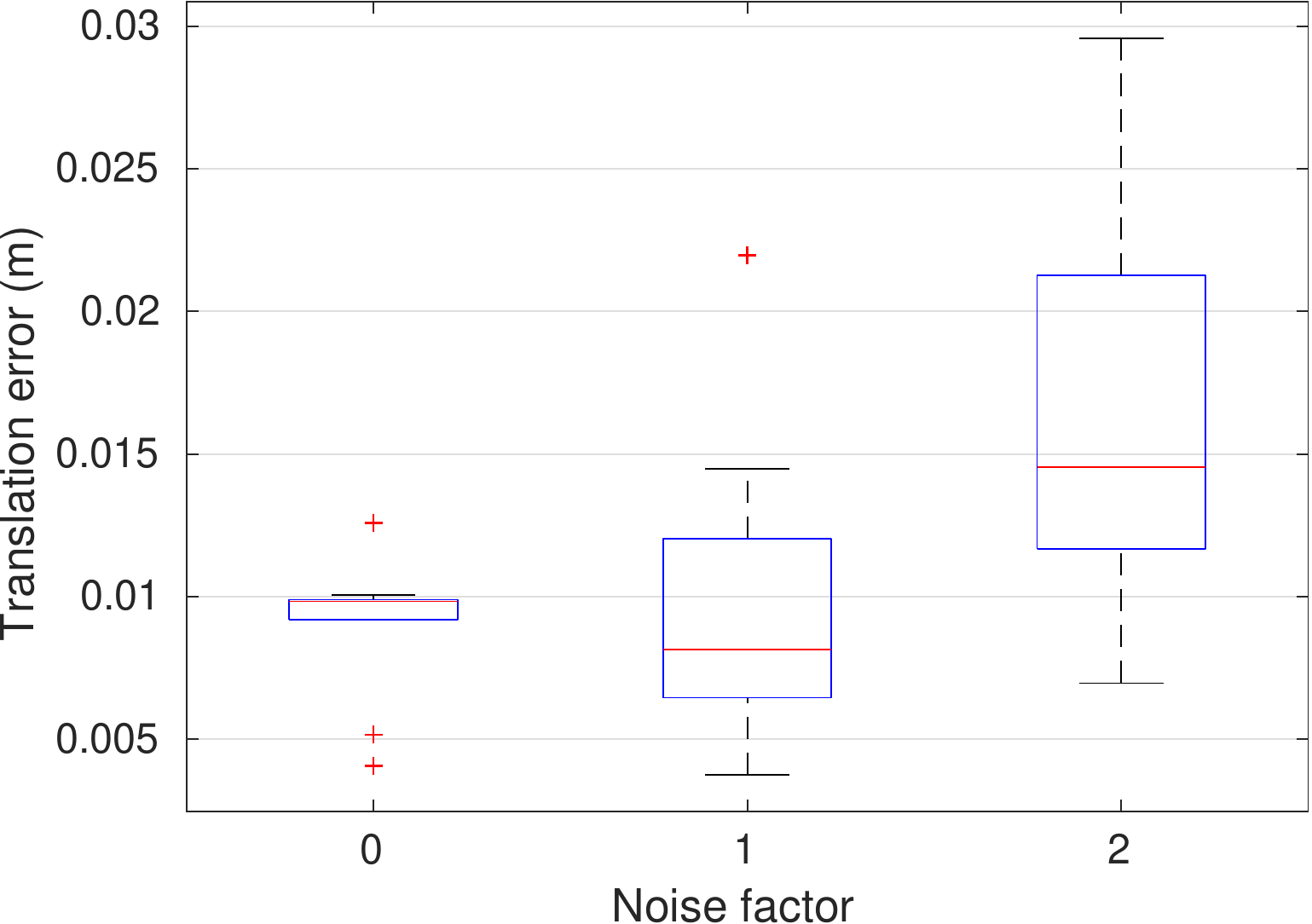}%
\label{fig:rob_t}}
\hfil
\subfloat[Rotation error, $e_r$]{\includegraphics[height=1.12in]{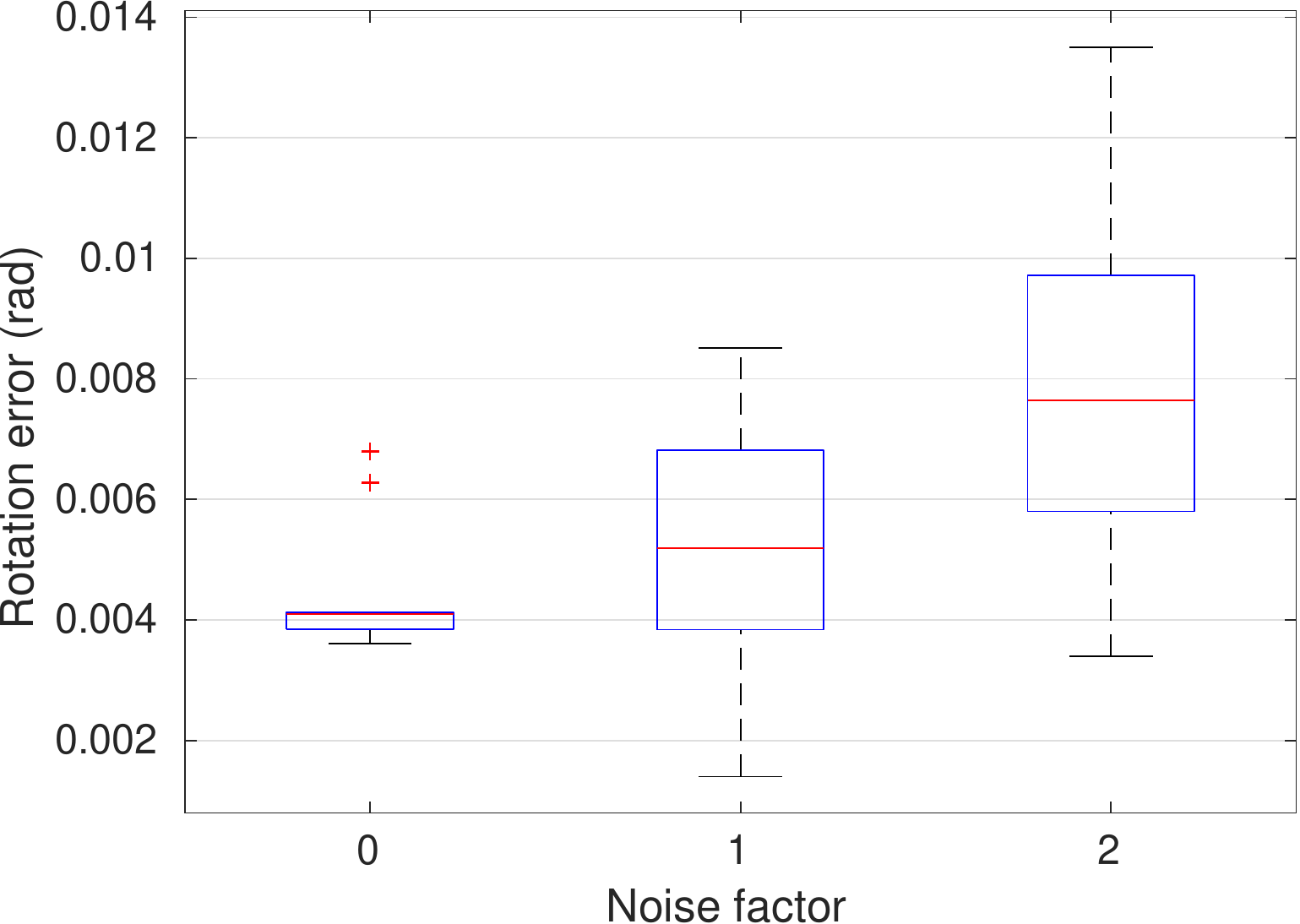}%
\label{fig:rob_a}}
\caption{Robustness of the calibration method against Gaussian noise in the data. The red line in the boxes is the median value.}
\label{fig:rob}
\end{figure}

On the other hand, we tested the robustness of the method against different levels of noise in the input data. Modeling the noise with a normal distribution $\mathcal{N}(0, (K \sigma_0)^2)$, we provide the calibration error for three different values of the \textit{noise factor} $K$ in Fig. \ref{fig:rob}.

\begin{figure}[htb]
\centering
\subfloat[]{\includegraphics[height=1in]{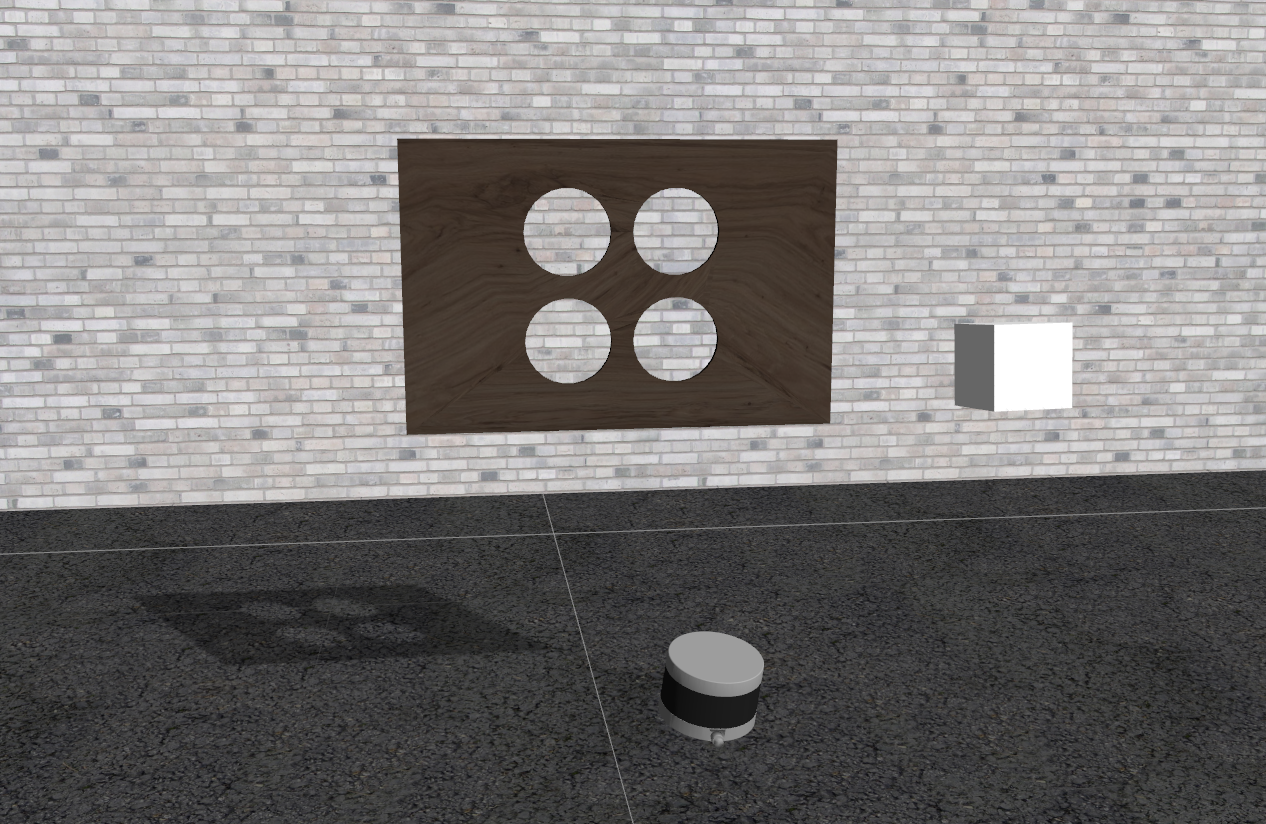}%
\label{fig:patronworld}}
\hfil
\subfloat[]{\includegraphics[height=1in]{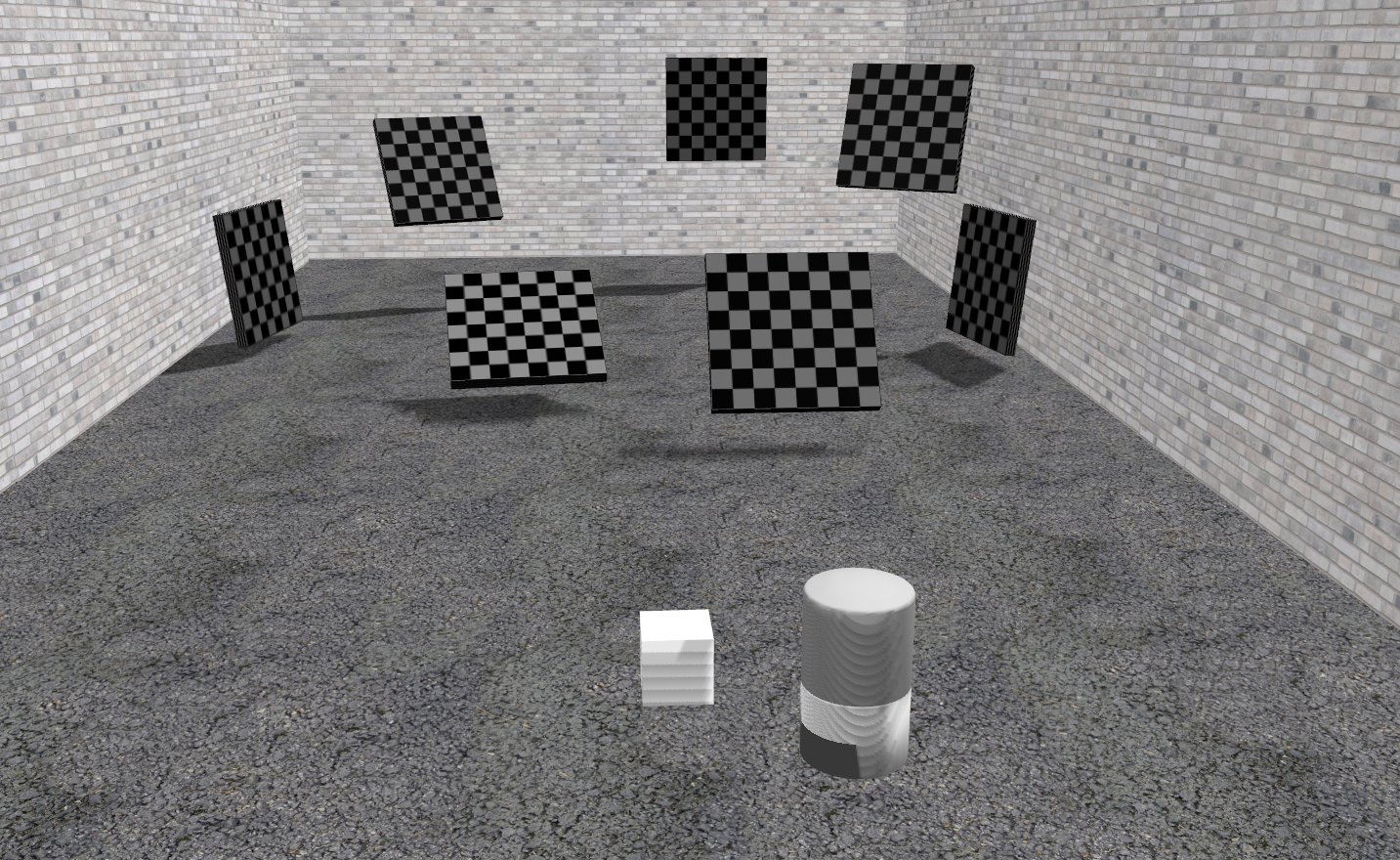}%
\label{fig:geigerworld}}
\caption{Simulator scenes: (a) custom calibration target; (b) Geiger et al. setup.}
\label{fig:worlds} 
\end{figure}

We also compared our method with those proposed by Velas et. al \cite{Velas2014}, using their ROS implementation, and Geiger et al. \cite{Geiger2012c}, through their public web toolbox\footnote{http://www.cvlibs.net/software/calibration/}. For fairness, it is important to note that both methods are aimed to monocular cameras and, in addition, the latter one is able to provide the camera intrinsic parameters as well. Experiments for Geiger et. al were conducted on a representative set of settings (1, 2, 4, 8 and 9) and the best result was chosen. Meanwhile, Velas et al. was applied over the whole set of settings. Nevertheless, it was unable to provide valid results for most of the settings since it is designed for smaller magnitudes of the transformation parameters.
The required scenarios for the different calibration methods were properly recreated in the simulator, as shown in Fig. \ref{fig:worlds}.

\begin{figure}[htb]
\centering
\subfloat[16-layer, trans. error]{\includegraphics[height=1.1in]{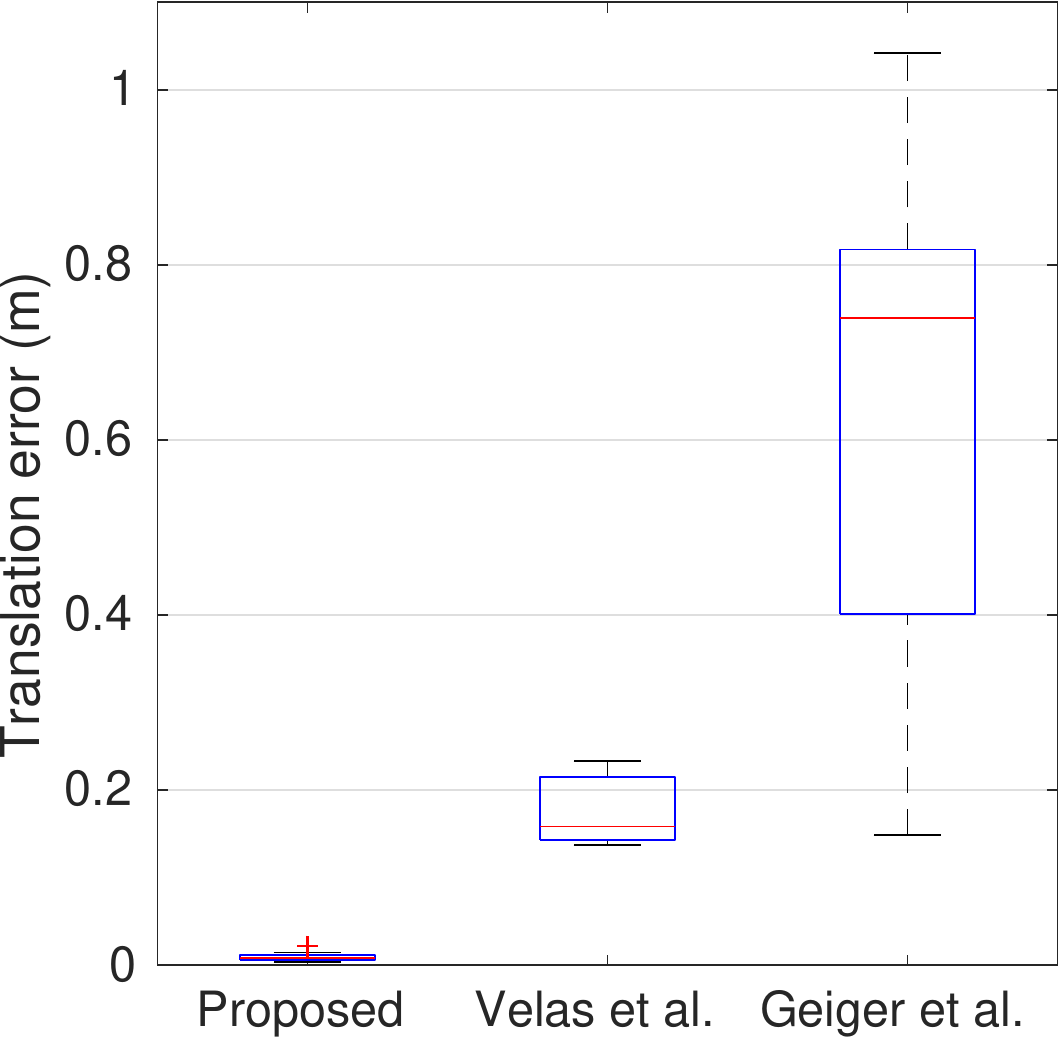}%
\label{fig:comp_t1}}
\hfill
\subfloat[32-layer, trans. error]{\includegraphics[height=1.1in]{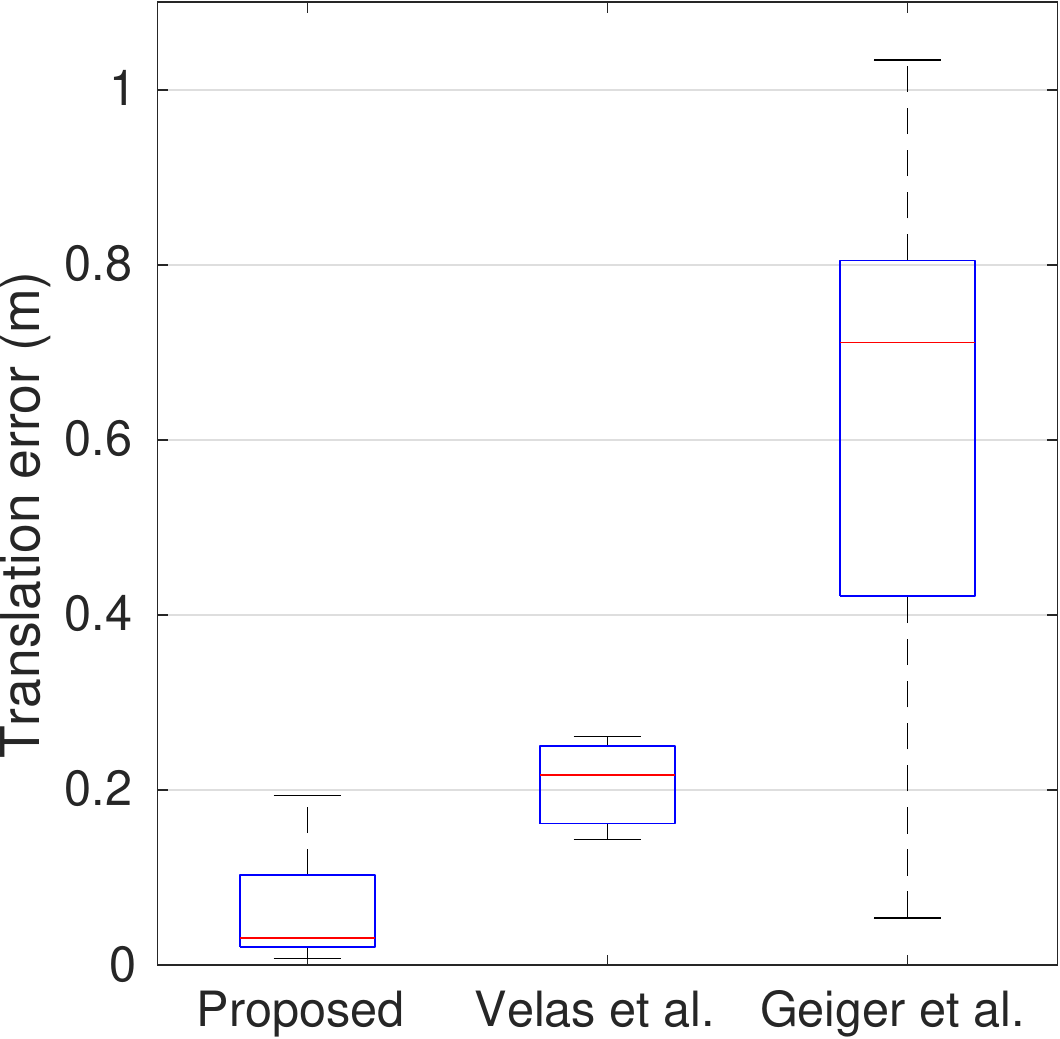}%
\label{fig:comp_t2}}
\hfill
\subfloat[64-layer, trans. error]{\includegraphics[height=1.1in]{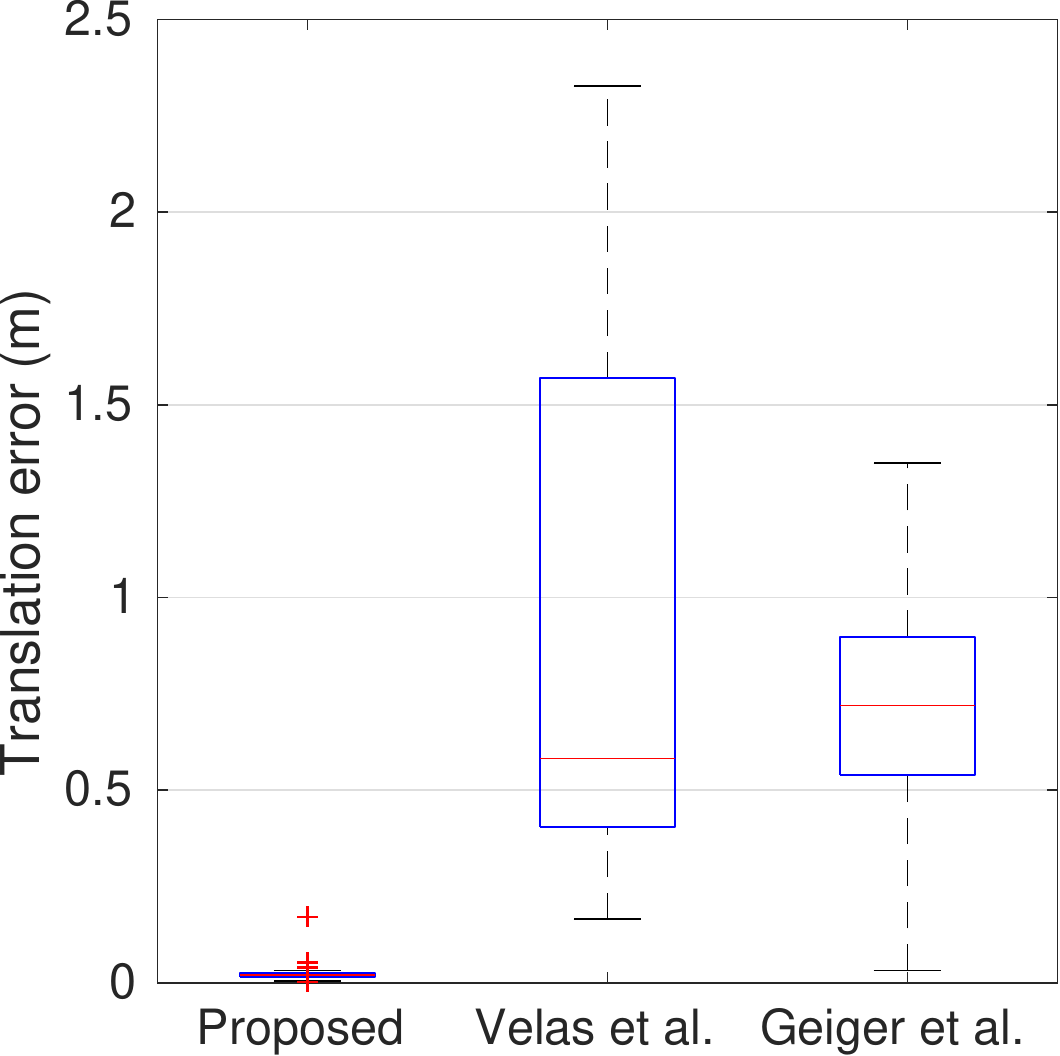}%
\label{fig:comp_t3}}
\\
\subfloat[16-layer, rot. error]{\includegraphics[height=1.1in]{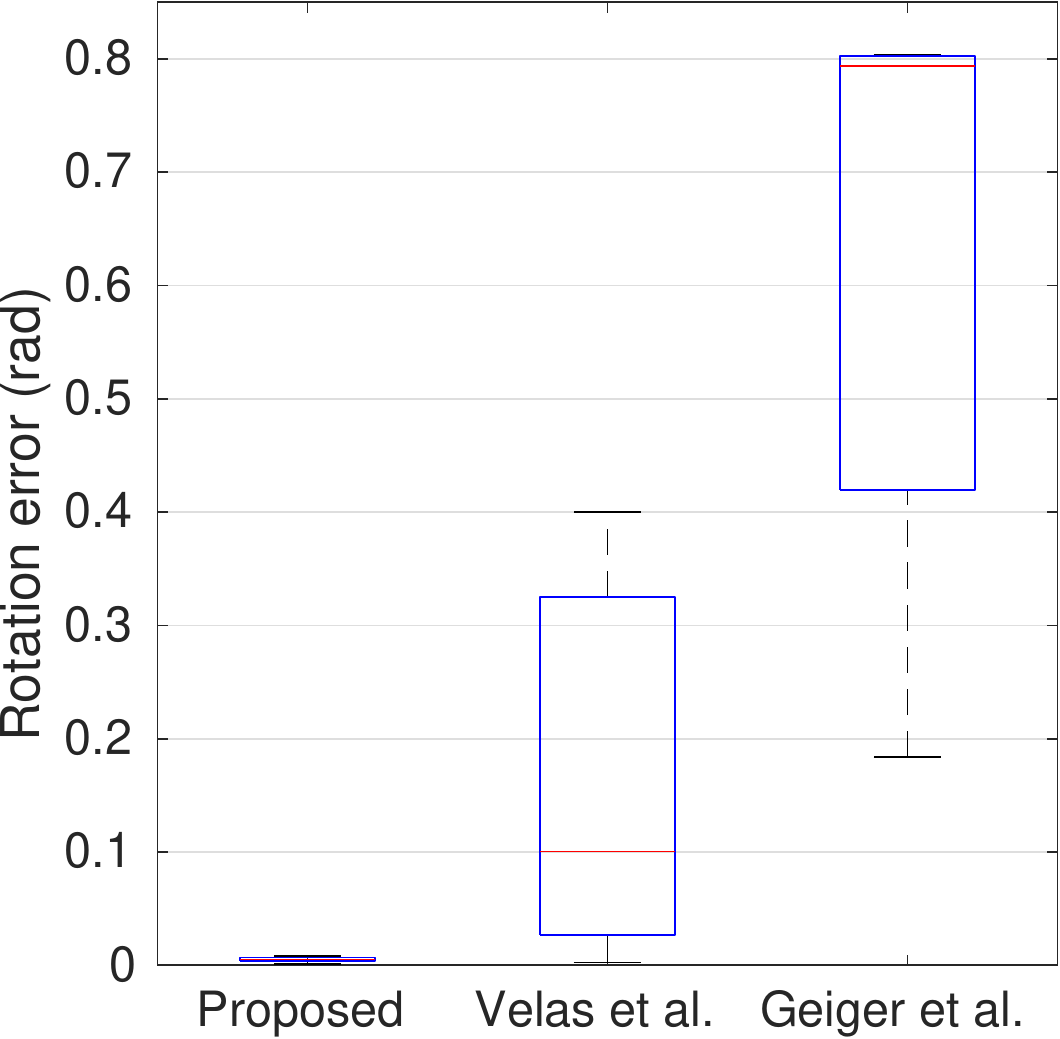}%
\label{fig:comp_a1}}
\hfill
\subfloat[32-layer, rot. error]{\includegraphics[height=1.1in]{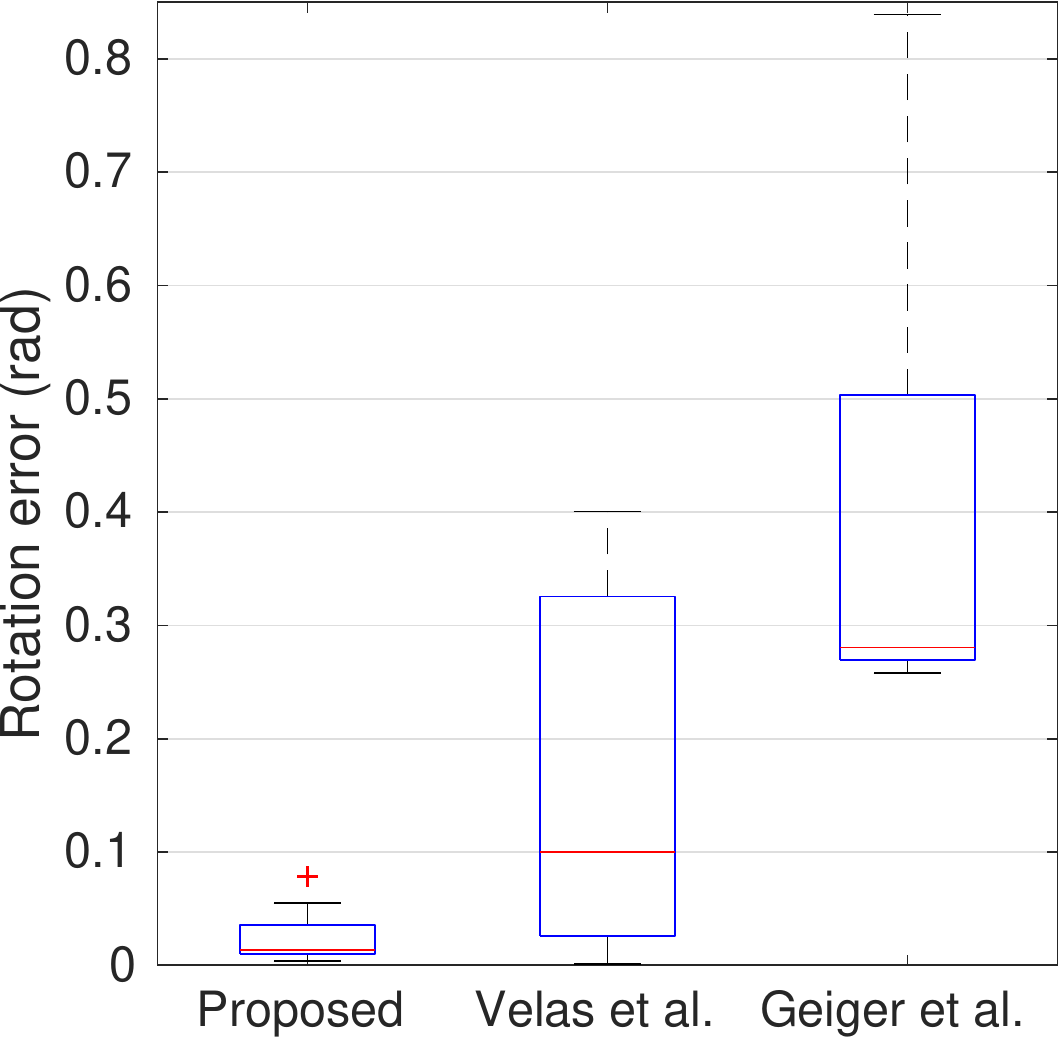}%
\label{fig:comp_a2}}
\hfill
\subfloat[64-layer, rot. error]{\includegraphics[height=1.1in]{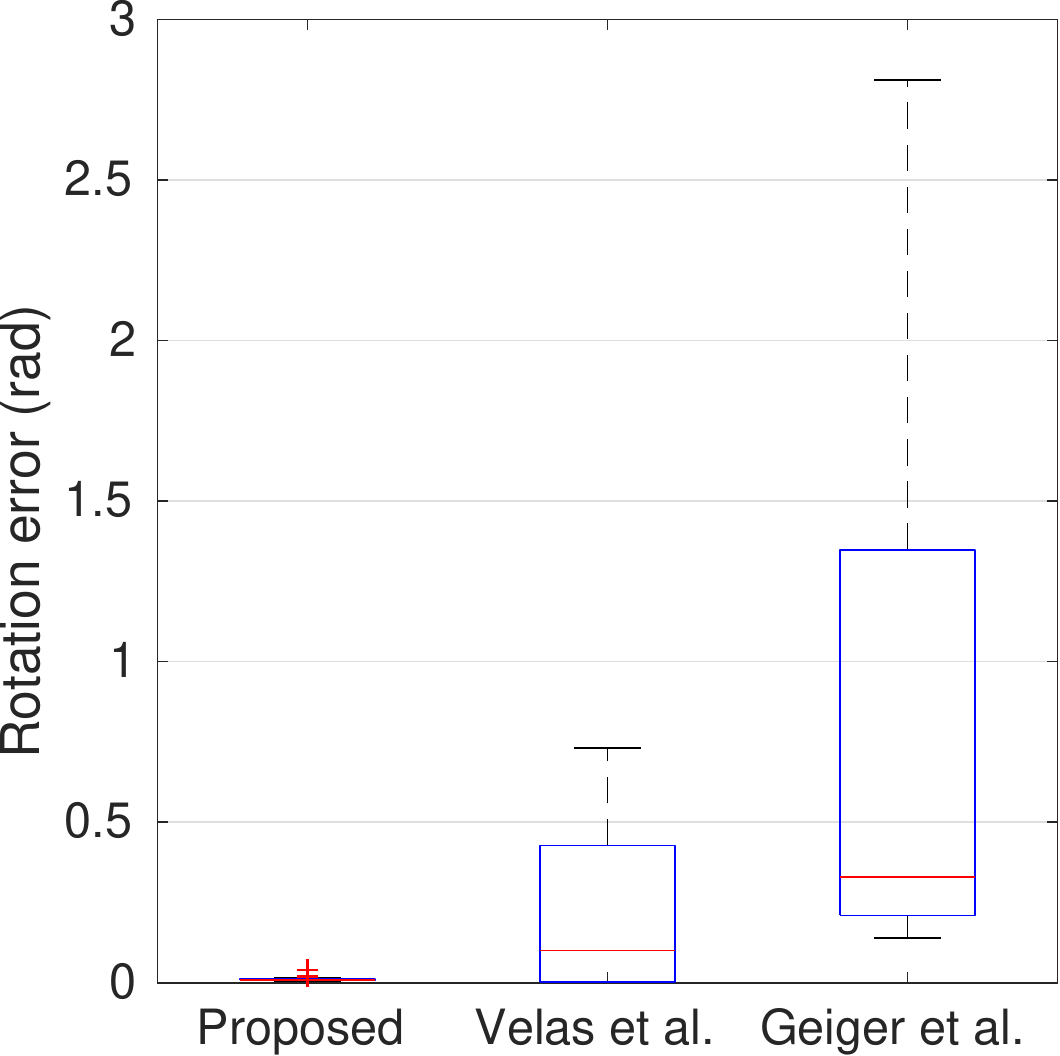}%
\label{fig:comp_a3}}
\caption{Comparison of reprojection errors with other comparable methods.}
\label{fig:comp}
\end{figure}

Since the works we used for comparison were not targeted for mid-resolution laser, the performance evaluation was conducted over three lidar devices with a different number of scan planes: 16, 32 and 64. Results, considering only valid outcomes, are presented in Fig. \ref{fig:comp}.

\subsection{Tests in Real Scenarios}
\begin{figure}[htb]
\centering
\subfloat[]{\includegraphics[height=1in]{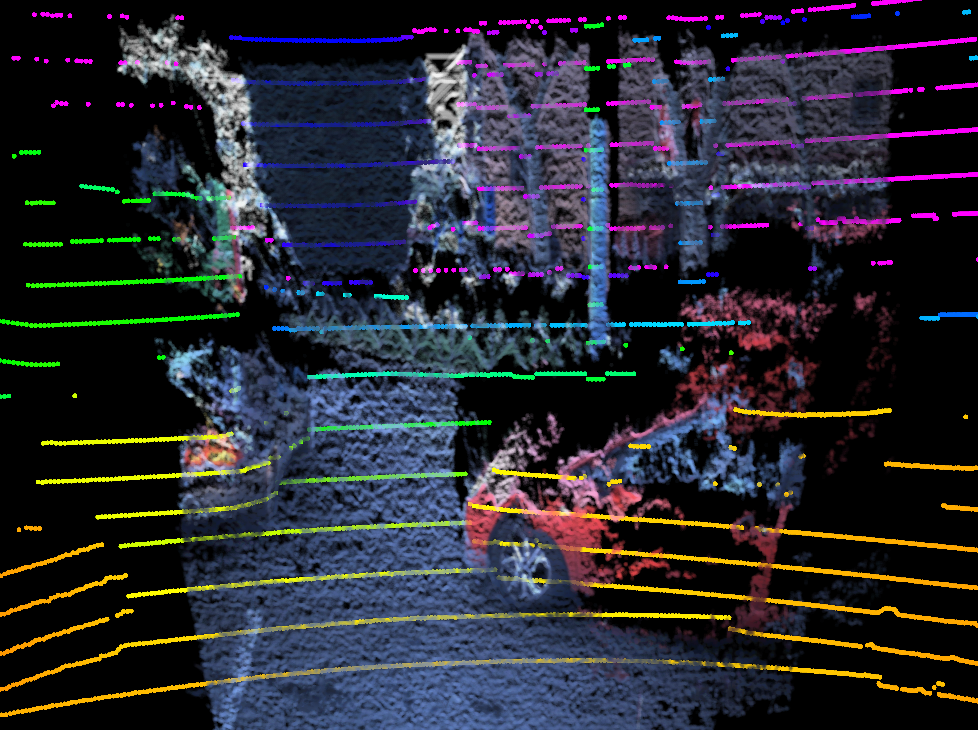}%
\label{fig:real_cloud}}
\hfil
\subfloat[]{\includegraphics[height=1in]{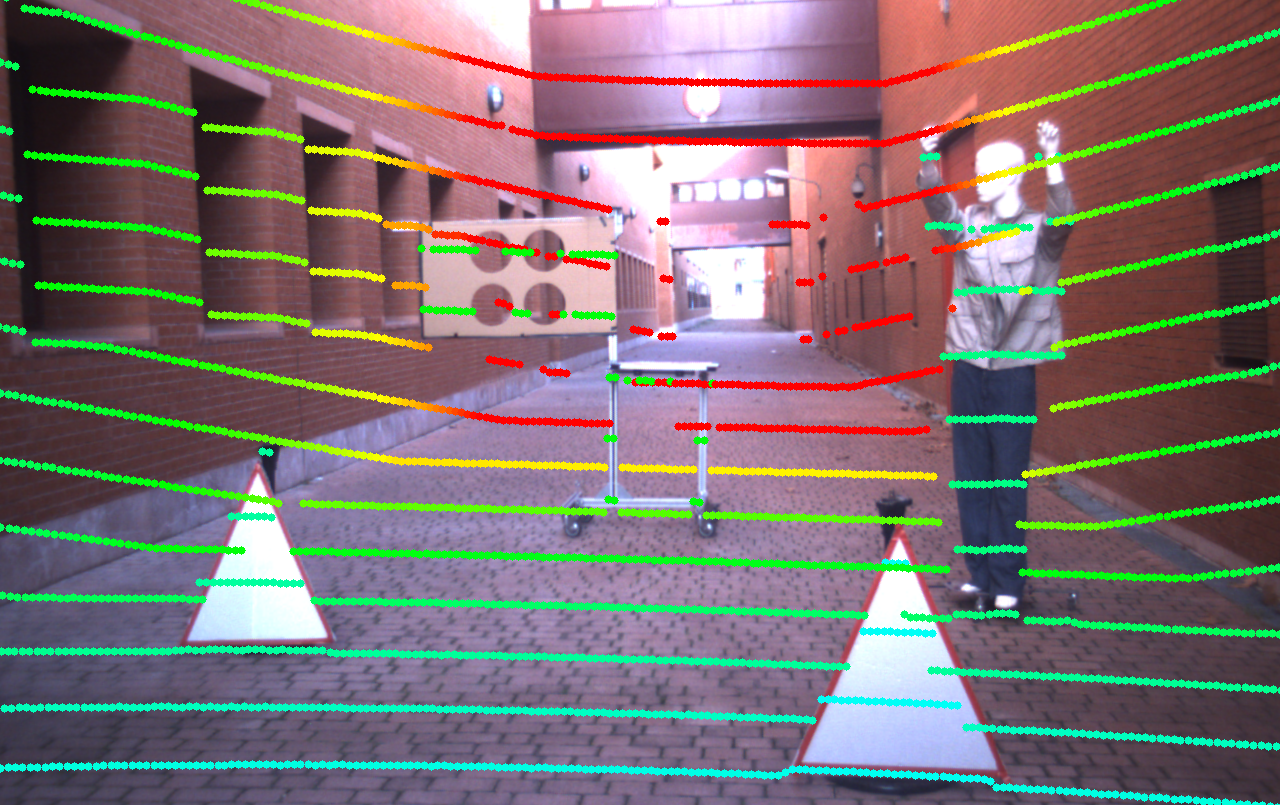}%
\label{fig:real_img}}
\caption{Results in real scenes: (a) cloud registration; (b) range projections.}
\label{fig:realtest}
\end{figure}

Tests in real scenarios were performed using the IVVI 2.0 platform setup. The results proved the usability of the algorithm for automotive applications requiring data registration, as shown in Fig. \ref{fig:realtest}.

\section{Conclusion}
\label{sec:conclusion}
We have presented a methodology for calibration of lidar-stereo camera systems aimed to remove nearly all the burden associated with the process. Contrary to previous approaches, we focus on getting accurate results in real setups using close-to-production sensor devices, without user intervention. 

We have also addressed the open issue of the assessment of the calibration performance, caused by the inability of obtaining precise ground truth measurements of the sensor setups in practice. An advanced simulation software has been used to accurately recreate the sensor models and the calibration scenes, with the inherent advantage of having the exact measurements available for the validation of the methods. 

Experiments show that the proposed algorithm outperforms the existing approaches by a large margin, while real tests have corroborated the results provided by the simulation software.

For the sake of research reproducibility, we have released our open-source implementation of the test suite, as well as the calibration algorithm, both integrated into the ROS framework. 

\section*{Acknowledgement} 
Research supported by the Spanish Government through the CICYT projects (TRA2015-63708-R and TRA2016-78886-C3-1-R), and the Comunidad de Madrid through SEGVAUTO-TRIES (S2013/MIT-2713).

\bibliographystyle{IEEEtran}
\bibliography{IEEEabrv,carlos}

\begin{thebibliography}{10}
\providecommand{\url}[1]{#1}
\csname url@samestyle\endcsname
\providecommand{\newblock}{\relax}
\providecommand{\bibinfo}[2]{#2}
\providecommand{\BIBentrySTDinterwordspacing}{\spaceskip=0pt\relax}
\providecommand{\BIBentryALTinterwordstretchfactor}{4}
\providecommand{\BIBentryALTinterwordspacing}{\spaceskip=\fontdimen2\font plus
\BIBentryALTinterwordstretchfactor\fontdimen3\font minus
  \fontdimen4\font\relax}
\providecommand{\BIBforeignlanguage}[2]{{%
\expandafter\ifx\csname l@#1\endcsname\relax
\typeout{** WARNING: IEEEtran.bst: No hyphenation pattern has been}%
\typeout{** loaded for the language `#1'. Using the pattern for}%
\typeout{** the default language instead.}%
\else
\language=\csname l@#1\endcsname
\fi
#2}}
\providecommand{\BIBdecl}{\relax}
\BIBdecl

\bibitem{Franke2013}
U.~Franke, D.~Pfeiffer, C.~Rabe, C.~Knoeppel, M.~Enzweiler, F.~Stein, and R.~G.
  Herrtwich, ``{Making Bertha See},'' in \emph{IEEE International Conference on
  Computer Vision Workshops (ICCVW)}, dec 2013, pp. 214--221.

\bibitem{Garcia2017}
F.~Garc{\'{i}}a, D.~Mart{\'{i}}n, A.~de~la Escalera, and J.~M. Armingol,
  ``{Sensor Fusion Methodology for Vehicle Detection},'' \emph{IEEE Intelligent
  Transportation Systems Magazine}, vol.~9, no.~1, pp. 123--133, 2017.

\bibitem{Kwak2011}
K.~Kwak, D.~F. Huber, H.~Badino, and T.~Kanade, ``{Extrinsic Calibration of a
  Single Line Scanning Lidar and a Camera},'' in \emph{Proc. IEEE/RSJ
  International Conference on Intelligent Robots and Systems (IROS)}, 2011, pp.
  3283--3289.

\bibitem{Scaramuzza2007}
D.~Scaramuzza, A.~Harati, and R.~Siegwart, ``{Extrinsic Self Calibration of a
  Camera and a 3D Laser Range Finder from Natural Scenes},'' in \emph{Proc.
  IEEE/RSJ International Conference on Intelligent Robots and Systems (IROS)},
  2007, pp. 4164--4169.

\bibitem{Debattisti2013}
S.~Debattisti, L.~Mazzei, and M.~Panciroli, ``{Automated Extrinsic Laser and
  Camera Inter-Calibration Using Triangular Targets},'' in \emph{Proc. IEEE
  Intelligent Vehicles Symposium (IV)}, 2013, pp. 696--701.

\bibitem{Park2014}
Y.~Park, S.~Yun, C.~S. Won, K.~Cho, K.~Um, and S.~Sim, ``{Calibration between
  Color Camera and 3D LIDAR Instruments with a Polygonal Planar Board},''
  \emph{Sensors}, vol.~14, no.~3, pp. 5333--5353, 2014.

\bibitem{Pereira2016}
M.~Pereira, D.~Silva, V.~Santos, and P.~Dias, ``{Self calibration of multiple
  LIDARs and cameras on autonomous vehicles},'' \emph{Robotics and Autonomous
  Systems}, vol.~83, pp. 326--337, 2016.

\bibitem{Li2011}
Y.~Li, Y.~Ruichek, and C.~Cappelle, ``{3D Triangulation Based Extrinsic
  Calibration between a Stereo Vision System and a LIDAR},'' in \emph{Proc.
  IEEE International Conference on Intelligent Transportation Systems (ITSC)},
  2011, pp. 797--802.

\bibitem{Scott2015}
T.~Scott, A.~A. Morye, P.~Pini{\'{e}}s, L.~M. Paz, I.~Posner, and P.~Newman,
  ``{Exploiting Known Unknowns: Scene Induced Cross-Calibration of Lidar-Stereo
  Systems},'' in \emph{Proc. IEEE/RSJ International Conference on Intelligent
  Robots and Systems (IROS)}, 2015, pp. 3647--3653.

\bibitem{Geiger2012c}
A.~Geiger, F.~Moosmann, {\"{O}}.~Car, and B.~Schuster, ``{Automatic Camera and
  Range Sensor Calibration using a single Shot},'' in \emph{Proc. IEEE
  International Conference on Robotics and Automation (ICRA)}, 2012, pp.
  3936--3943.

\bibitem{Velas2014}
M.~Velas, M.~Spanel, Z.~Materna, and A.~Herout, ``{Calibration of RGB Camera
  With Velodyne LiDAR},'' in \emph{Comm. Papers Proc. International Conference
  on Computer Graphics, Visualization and Computer Vision (WSCG)}, 2014, pp.
  135--144.

\bibitem{Moghadam2013}
P.~Moghadam, M.~Bosse, and R.~Zlot, ``{Line-based Extrinsic Calibration of
  Range and Image Sensors},'' in \emph{Proc. IEEE International Conference on
  Robotics and Automation (ICRA)}, 2013, pp. 3685--3691.

\bibitem{Ponz}
C.~H. {Rodr{\'{i}}guez Garavito}, A.~Ponz, F.~Garc{\'{i}}a, D.~Mart{\'{i}}n,
  A.~de~la Escalera, and J.~M. Armingol, ``{Automatic Laser And Camera
  Extrinsic Calibration for Data Fusion Using Road Plane},'' in \emph{IEEE
  International Conference on Information Fusion (FUSION)}, 2014.

\bibitem{Levinson2013}
J.~Levinson and S.~Thrun, ``{Automatic Online Calibration of Cameras and
  Lasers},'' in \emph{Robotics: Science and Systems}, 2013.

\bibitem{Hirschmuller2008}
H.~Hirschm{\"{u}}ller, ``{Stereo Processing by Semiglobal Matching and Mutual
  Information},'' \emph{IEEE Transactions on Pattern Analysis and Machine
  Intelligence}, vol.~30, no.~2, pp. 328--341, 2008.

\bibitem{Besl1992}
P.~J. Besl and N.~D. McKay, ``{A method for Registration of 3-D Shapes},''
  \emph{IEEE Transactions on Pattern Analysis and Machine Intelligence},
  vol.~14, no.~2, pp. 239--256, 1992.

\bibitem{Martin2014a}
D.~Mart{\'{i}}n, F.~Garc{\'{i}}a, B.~Musleh, D.~Olmeda, G.~A. Pel{\'{a}}ez,
  P.~Mar{\'{i}}n, A.~Ponz, C.~H. {Rodr{\'{i}}guez Garavito}, A.~Al-Kaff,
  A.~de~la Escalera, and J.~M. Armingol, ``{IVVI 2.0: An intelligent vehicle
  based on computational perception},'' \emph{Expert Systems with
  Applications}, vol.~41, no.~17, pp. 7927--7944, dec 2014.

\bibitem{Koenig2004}
N.~Koenig and A.~Howard, ``{Design and Use Paradigms for Gazebo, An Open-Source
  Multi-Robot Simulator},'' in \emph{Proc. IEEE/RSJ International Conference on
  Intelligent Robots and Systems (IROS)}, 2004, pp. 2149--2154.

\end{thebibliography}

\end{document}